\documentclass{article}

\PassOptionsToPackage{numbers, compress}{natbib}

\usepackage{algorithm}
\usepackage{algpseudocode}
\usepackage[main, final]{neurips_2026}

\usepackage[utf8]{inputenc} 
\usepackage[T1]{fontenc}    
\usepackage{hyperref}       
\usepackage{url}            
\usepackage{booktabs}       
\usepackage{amsfonts}       
\usepackage{nicefrac}       
\usepackage{microtype}      
\usepackage{xcolor}         
\usepackage{graphicx}
\usepackage{subcaption}
\usepackage{enumitem}

\definecolor{CodeBlue}{RGB}{237,0,140}

\usepackage{amsmath}
\usepackage{amssymb}
\usepackage{mathtools}
\usepackage{amsthm}
\usepackage{multirow}
\usepackage[table]{xcolor}  
\usepackage{pifont}
\newcommand{\cmark}{\ding{51}} 
\newcommand{\xmark}{\ding{55}} 

\definecolor{headergray}{RGB}{220,220,220}
\definecolor{bestred}{RGB}{245,175,175}        
\definecolor{secondorange}{RGB}{245,215,170}   
\newcommand{\best}[1]{\cellcolor{bestred}#1}
\newcommand{\second}[1]{\cellcolor{secondorange}#1}

\usepackage[capitalize,noabbrev]{cleveref}
\theoremstyle{plain}

\theoremstyle{definition}

\theoremstyle{remark}

\usepackage[textsize=tiny]{todonotes}
\usepackage{wrapfig}

\title{PhyMo: Learning Physical Dynamics with Accurate and Continuous Motion from
Multi-View Videos}

\author{%
  Shangjia Liu$^{\S}$ \quad
  Jinxi Li$^{\dagger}$ \quad
  Siyuan Zhou$^{\dagger}$ \quad
  Bo Yang$^{\dagger}$ \quad
  Bing Wang$^{\S}$\thanks{Corresponding Author.} \\[2pt]
  \\
  $^{\S}$Spatial Intelligence Group \qquad
  $^{\dagger}$vLAR Group \\
  The Hong Kong Polytechnic University, Hong Kong SAR, China \\
  \texttt{shangjia.liu@connect.polyu.hk} \quad
  \texttt{bingwang@polyu.edu.hk}
}

\begin{document}

\maketitle

\begin{abstract}
 Recovering dynamic 3D scene geometry, appearance, and the physical motion states that govern scene evolution from multi-view videos is important yet challenging. A key difficulty is that appearance can be deceptive.
 Under rendering supervision alone, a model may fit the observed images by changing color or opacity to compensate for inaccurate deformation, rather than recovering the true evolving geometry and motion.
 This issue is especially severe in scenes with fast or locally complex motion, where inaccurate trajectories can still produce plausible images over the observed frames, but lead to unreliable reconstruction and discontinuous future prediction. In this paper, we propose \textbf{PhyMo}, a framework for learning evolving dynamics from multi-view videos. We introduce a {Motion-Aware Trajectory Representation} that drives the model to recover accurate motion from deformation, rather than relying on appearance shortcuts to explain observed motion. On top of this, we propose a {Continuous Motion Evolution} module with bidirectional kinematic continuity constraint to promote temporally consistent evolution of displacement, velocity, and acceleration. Together, these designs yield geometrically faithful interpolation and continuous motion extrapolation. Experiments on four dynamic scene benchmarks demonstrate state-of-the-art performance. Code is available at
{\hypersetup{pdfborder={0 0 0}}%
\color{CodeBlue}\url{https://github.com/BystanderJ/PhyMo}}.
\end{abstract}

\section{Introduction}
Modeling the real, complex, and dynamic physical 3D world has long been a fundamental yet challenging problem in computer vision and robotics. A faithful understanding of how 3D scenes evolve over time is critical for a wide range of downstream applications, including embodied AI, world models, and robotic motion planning. Recent advances in 3D scene representations, such as Neural Radiance Fields\cite{mildenhall2021nerf} and 3D Gaussian Splatting\cite{kerbl20233d}, have enabled high-quality dynamic scene reconstruction from visual observations. With such powerful representations, current research largely centers on reconstruction quality, yet two fundamental questions remain: what constitutes dynamics in a scene representation, and how can we learn it effectively?

One line of work~\cite{wan2024superpoint,cao2023hexplane,yang2024deformable,kwak2025modec} views dynamics as a spatial-temporal trajectory and represents motion with a deformation field. These methods typically fit trajectories by uniformly sampling timestamps and jointly optimizing the deformation and canonical appearance to match observations. However, since their motion modeling is not grounded in physics, they mainly interpolate within the observed interval and degrade quickly beyond it.
A second line of work~\cite{chu2022physics,li2023nvfi,li2023pac,cho2024parameterized} models dynamics as physical motion states, such as velocity (and sometimes higher-order states), and learns them via integration to match observations. While this formulation enables future extrapolation, integration-based learning is prone to error accumulation, which can compromise reconstruction quality.
More recently, hybrid approaches combine both representations~\cite{zhao2024gaussianprediction,li2025freegave,caleb20254d,li2025trace,wang2025ode}, using spatial-temporal trajectories to guide the learning of physical motion states and aiming to better balance reconstruction fidelity and future prediction. However, these methods often rely primarily on photometric supervision, leaving motion states weakly constrained and lacking temporal consistency. This limitation reflects a broader gap shared by existing approaches, regardless of the specific dynamics representation.

In this paper, we aim to recover physically plausible scene evolution from dynamic scenes, rather than merely reconstructing geometry and appearance, so that the learned representation supports both {geometrically faithful interpolation} and {continuous motion extrapolation}. However, this task is extremely challenging, because existing methods have not resolved two key issues: 
\emph{1) Rendering fidelity is not equivalent to faithful motion}. Although motion may remain smooth for long periods, recovering accurate trajectories becomes much harder during rapid changes or state transitions. Existing methods typically optimize deformation and canonical appearance jointly under rendering supervision, creating a fundamental ambiguity: while accurate motion can explain correct appearance, correct appearance alone is not evidence of accurate motion. As a result, they may produce plausible renderings while still recovering unreliable trajectories and deformations.
\emph{2) Motion evolution is inherently continuous}. Displacement, velocity, and acceleration should propagate consistently over time according to the underlying kinematic relationships. However, existing methods often rely on learned trajectories and rendering supervision alone to recover these motion states, without explicitly constraining their temporal consistency. Consequently, the estimated motion states may be locally  discontinuous over time, leading to unreliable motion details in both interpolation and extrapolation.

\begin{wrapfigure}{r}{0.58\linewidth}
    \centering
    \vspace{-10pt}
    \includegraphics[width=1.0\linewidth]{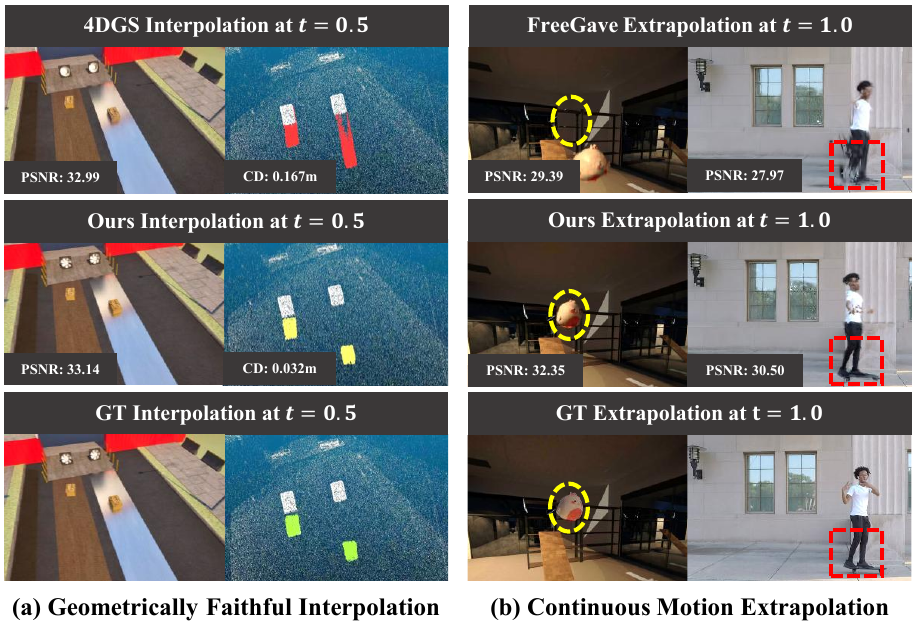}
    \caption{
Given video of a dynamic scene, our PhyMo can achieve geometrically faithful reconstruction and continuous motion prediction, comparing to other methods. In (a), white points denote the canonical point cloud, while colored points denote the dynamic point cloud at $t = 0.5$.}
    \label{qua_res}
    \vspace{-10pt}
\end{wrapfigure}

To address these challenges, we propose \textbf{PhyMo}, a framework for learning \textbf{phy}sical evolution with accurate and continuous \textbf{mo}tion  from multi-view videos.
Specifically, we introduce a \emph{Motion-aware Trajectory Representation} that drives the model to recover accurate motion through deformation, so that observed dynamics are explained by deformation rather than by shortcuts in canonical appearance. This mitigates a key failure mode of existing methods, where fast motion or locally complex motion can be visually explained by appearance fitting while the underlying deformation is still wrong. As a result, our method recovers dynamic scenes that are not only visually plausible, but also \emph{geometrically faithful} in 3D space. On top of the recovered motion trajectories, we further introduce a \emph{Continuous Motion Evolution} module to model the evolution of physical motion states. Since estimating displacement, velocity, and acceleration from discrete observations under rendering supervision alone is inherently under-constrained, we impose a bidirectional kinematic continuity constraint to promote temporally continuous state propagation, thereby enabling  \emph{continuous motion extrapolation}.

Overall, our method addresses the under-constrained problem of recovering physical motion states like displacement, velocity, acceleration from visual supervision alone. By combining motion-aware trajectory representation with continuous motion evolution, PhyMo achieves both geometrically faithful interpolation and continuous motion extrapolation.
Figure~\ref{qua_res} shows our qualitative results on two representative tasks, our contributions can be summarized as follows:

\begin{itemize}[leftmargin=*,itemsep=0.1em,topsep=0.05em,parsep=0.1pt,partopsep=0pt]
\item We introduce a new framework for physically plausible dynamic scene modeling from multi-view videos. By combining Motion-Aware Trajectory Representation with Continuous Motion Evolution, PhyMo achieves geometrically faithful reconstruction and continuous motion prediction.
\item We demonstrate the \textbf{appearance shortcut} problem in dynamic reconstruction and mitigate it with a {Motion-Aware Trajectory Representation}, which drives dynamics  to be captured by deformation rather than appearance shortcuts for geometrically faithful reconstruction.
\item We propose a {Continuous Motion Evolution} module with bidirectional kinematic continuity to enforce temporally consistent displacement, velocity, and acceleration, and demonstrate strong performance on both  interpolation and extrapolation across four benchmarks.
\end{itemize}

\section{Related Works}
\textbf{3D Representations.}
Traditional static 3D representations, such as meshes\cite{kato2018neural}, point clouds\cite{qi2017pointnet}, voxel grids\cite{yang2018dense}, and primitives\cite{zou20173d}, are explicit yet inherently discrete. As a result, they often suffer from fundamental limitations in fidelity and efficiency. Recently, a series of implicit 3D modeling approaches have been proposed, such as occupancy fields\cite{chen2019learning,mescheder2019occupancy}, signed/unsigned distance fields\cite{deng2026primitiveudf,chibane2020neural,wang2022rangeudf,park2019deepsdf}, and radiance fields\cite{mildenhall2021nerf}. Although these methods have significantly advanced novel-view synthesis and shape reconstruction, their rendering is often time-consuming due to expensive volumetric sampling. To address this issue, 3D Gaussian Splatting (3DGS)\cite{kerbl20233d} and its variants\cite{fang2024mini,park2021nerfies,fang2025efficient,lei2025mosca,gu2025mmone} adopts a point-based representation to encode 3D properties, enabling faster rendering. In our framework, we choose 3DGS as the scene representation, since it naturally integrates with  of our pipeline.

\textbf{Dynamic 3D Reconstruction.}
Recent progress in dynamic 3D reconstruction has largely focused on extending static scene representations, such as NeRF and 3D Gaussian Splatting (3DGS), into the spatio-temporal domain~\cite{gan2023v4d,katsumata2024compact,cai2022neural,cai2022neural,gao2021dynamic,wang2022dm,lin2024gaussian,liu2023robust,tretschk2021non,morkva2026mosaic,huang2026mango}. 
Most NeRF-based or 3DGS-based approaches decouple the static representation from motion or deformation, and incorporate time as an additional input to learn a deformation or motion field. 
While these methods achieve impressive visual quality for appearance reconstruction, their learned dynamics are often weakly constrained by rendering supervision, limiting their reliability for both past retrodiction and future extrapolation. 
On the one hand, the learned temporal field may fit photometric correspondences without capturing physically meaningful motion evolution, leading to poor prediction beyond observed frames. 
On the other hand, even within the observed time span, visually accurate rendering does not necessarily imply accurate dynamic geometry or deformation. 
Recent efforts~\cite{wang2025freetimegs,liao2026sharptimegs} improve dynamic Gaussian representations by relaxing the canonical-space assumption, introducing short-range motion, or modeling temporal visibility and lifespan. 
However, these works mainly treat such failures as  optimization issues, without explicitly revealing the underlying competition between appearance fitting and motion recovery.

\textbf{3D Physics Learning.}
Motivated by the classical Physics-Informed Neural Networks (PINNs) framework, a growing body of research has explored learning physical attributes of 3D scenes and dynamic objects, such as Young's modulus and viscosity ~\cite{baieri2023fluid,im2024regularizing,raissi2020hidden,wang2024physics}. While these methods can enable future prediction, they are often constrained to a specific set of physical laws and governing equations. However, real-world dynamics are highly complex and are frequently difficult to capture with accurate simulation-based models. 
Another class of methods instead directly embeds physics systems~\cite{lin2025omniphysgs,xie2024physgaussian,sanchez2020learning,liu2024physics3d,mani2023surfsup,li2026learning,gu2026r}, such as graph-based interactions or mass spring structures into the network, so that the dynamics can be learned in a more structured manner.
However, such methods also tend to be limited in generalizability and scalability, since the embedded physics systems are usually tailored to specific object categories, material behaviors, or interaction patterns.
More recent hybrid methods~\cite{li2025freegave,li2025trace,wang2025ode} attempt to learn general physical motion states, such as velocity or acceleration, directly from dynamic videos. 
However, their motion-state learning is still largely dominated by rendering supervision, making the recovered physical states under-constrained: different motions may explain the observed frames similarly but yield unstable future predictions. 

\vspace{-1.0em}

\section{Method}

\begin{figure}[t]
  \begin{center}
    \centerline{\includegraphics[width=5.6in]{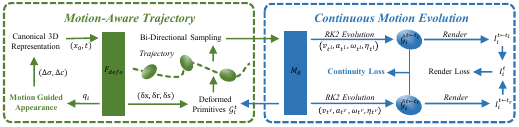}}
   \caption{\textbf{Overview of PhyMo.}
The left module, \textbf{Motion-Aware Trajectory }, starts from a canonical 3D representation and uses deformation network $F_{defo}$ to derive a motion-confidence gate $q_i$ that guides canonical appearance learning, encouraging motion to be explained by deformation rather than appearance shortcuts. The right module, \textbf{Continuous Motion Evolution}, bi-directionally sampling from the trajectory, Motion Network $M_{\theta}$ predicts velocity and acceleration and performs RK2 evolution , where bidirectional kinematic continuity loss and render loss are applied. }
    \label{overview}
  \end{center}
  \vspace{-1.8em}
\end{figure}

\vspace{-0.5em}
Given multi-view RGB videos of a dynamic scene, we denote the observed images as
$\{\mathbf{I}_t^n\}_{t=0,\,n=1}^{T,\,N}$, where $N$ is the total number of cameras and $T$ is the greatest timestamp in training.
Our goal is to recover faithful dynamic scene geometry and evolving motion from visual observations, enabling geometrically faithful  interpolation within the observed interval and continuous motion future-frame extrapolation beyond it.
As illustrated in Figure~\ref{overview}, our framework consists of two key components.
First, in Section~\ref{sec:motion_aware_traj}, we introduce a \textbf{Motion-Aware Trajectory Representation} that drives observed motion to be explained by deformation rather than by shortcuts in canonical appearance, thereby alleviating two common failure modes: premature appearance adaptation in fast local motion, and static-background appearance changes that compensate for unresolved motion.
Second, in Section \ref{2}, we introduce a \textbf{Continuous Motion Evolution} module that models the temporal evolution of displacement, velocity, and acceleration under bidirectional kinematic continuity, improving motion consistency over time.
Finally, in Section~\ref{3}, we describe the training strategy that optimizes the whole framework, allowing the learned representation to better capture complex dynamic scenes and produce reliable interpolation and extrapolation results.

\subsection{Motion-Aware Trajectory Representation}
\label{sec:motion_aware_traj}

\paragraph{Canonical 3D Gaussian Representation.}
We represent a dynamic scene by a set of canonical 3D Gaussian Splatting kernels, denoted as $\mathcal{G}_{0}$. Each canonical Gaussian is parameterized by its 3D mean position ${x}_{0}$, an anisotropic covariance matrix derived from a unit quaternion ${r}_{0}$ and per-axis scaling ${s}_{0}$, together with an opacity ${\sigma}$ and an SH coefficient vector ${c}$ for view-dependent appearance. Given the observed images at $t=0$, together with known camera poses and sparse Structure-from-Motion (SfM)~\cite{schonberger2016structure} points or random initialization, we optimize the canonical Gaussian set by the standard rendering-based learning procedure:
\begin{equation}
\underbrace{\big\{({x}_{0},{r}_{0},{s}_{0},\mathbf{\sigma},{c})\big\}}_{\mathcal{G}_{0}}
\;\xleftrightarrow[\;\;\ell_{1}+\ell_{\text{SSIM}}\;\;]{\text{project}+\text{render}}\;
\big\{\boldsymbol{I}_0^1,\ldots,\boldsymbol{I}_0^N\big\}
\label{eq:canonical_3dgs}
\end{equation}

\begin{figure*}[t]
  \begin{center}
    \centerline{\includegraphics[width=5.6in]{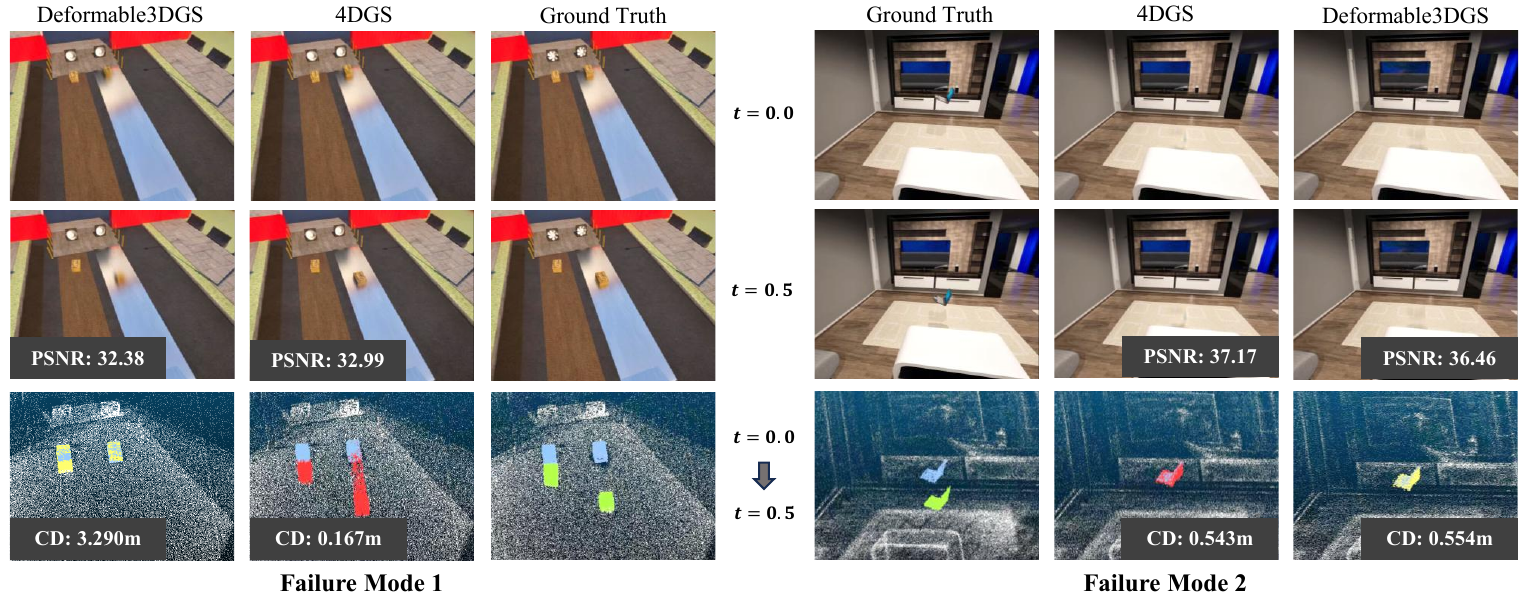}}
    \caption{
Two Failure Modes of 4D Reconstruction. \textbf{1. Static-background appearance shortcut:} Adjusting static-region appearance $c_i$ improves PSNR but doesn't recover evolving geometry. \textbf{2. Dynamic primitives disappearance shortcut:} For fast motion, the optimizer makes primitives transparent by reducing $\sigma_i$ instead of recovering the correct  trajectory, leading to a local minimum.
}
    \label{failuremode}
  \end{center}
  \vspace{-2.5em}
\end{figure*}

To model scene dynamics, we further introduce a deformation field $F_{\text{defo}}$ that predicts the time-dependent deformation of each Gaussian from its canonical position and time:
\begin{equation}
(\delta{x},\,\delta{r},\,\delta{s}) = F_{\text{defo}}({x}_{0},\,t)
\label{eq:defo}
\end{equation}
The Gaussian parameters at time $t$ are then obtained by
\begin{equation}
{x}_{t} = {x}_{0} + \delta{x},\quad
{r}_{t} = {r}_{0} \circ \delta{r},\quad
{s}_{t} = {s}_{0} \odot \delta{s},
\quad \mathbf{\sigma},\ {c}
\label{eq:deformed_gaussians}
\end{equation}
where $\circ$ and $\odot$ denote quaternion composition and element-wise scaling update, respectively. Although the deformation field is intended to model geometric changes over time, rendering-supervised optimization can reduce photometric error without learning reliable motion. 
Before the deformation field becomes accurate, the model may exploit appearance shortcuts through the canonical attributes ${c}$ and ${\sigma}$, leading to \textbf{two typical failure modes}, as shown in Figure~\ref{failuremode}. 
In static or background-dominated regions, adjusting the SH color coefficients ${c}$ can preserve visual quality even when the recovered dynamic geometry is inaccurate. 
In fast-moving or locally complex regions, reducing the opacity $\mathbf{\sigma}$ of poorly aligned dynamic Gaussians can make moving content fade or disappear instead of correcting its trajectory. 
Thus, rendering loss can decrease while motion recovery remains unreliable.

\textbf{Motion-Guided Appearance.} To suppress appearance shortcuts, we reparameterize the canonical appearance into a motion-guided residual form:
\begin{equation}
\mathbf{\sigma}_i = {\sigma}_i^{b} + q_i\,\Delta \mathbf{\sigma}_i,
\qquad
{c}_i = {c}_i^{b} + q_i\,\Delta {c}_i
\label{eq:motion_aware_app}
\end{equation}
where ${\sigma}_i^{b}$ and ${c}_i^{b}$ denote the base canonical opacity and SH coefficients, $\Delta \mathbf{\sigma}_i$ and $\Delta {c}_i$ are residual appearance parameters, and $q_i \in [0,1]$ is a per-Gaussian motion confidence gate. In this way, canonical appearance is no longer a fully free shortcut channel; instead, its effective update is guided by whether the corresponding Gaussian already exhibits active local motion.
To characterize whether the deformation field exhibits active local motion on a Gaussian, we define its motion confidence according to the deformation change between adjacent timestamps:
\begin{equation}
q_i =
\mathrm{sg}\!\left[
\psi\!\left(
\left\|\delta {x}_i(t_{k+1}) - \delta {x}_i(t_k)\right\|_2
\right)
\right]
\label{eq:motion_confidence}
\end{equation}
where $t_k$ and $t_{k+1}$ denote two adjacent sampled timestamps, $\psi(\cdot)$ is a monotonic squashing function mapping to $[0,1]$, and $\mathrm{sg}\left[\cdot\right]$ denotes stop-gradient. This design reflects whether the corresponding Gaussian exhibits active local motion, rather than how far it has moved away from the canonical state. Intuitively, if the deformation change between adjacent timestamps remains small, the corresponding Gaussian is still motion-inactive and its appearance residual should stay suppressed. Once a clear local motion response emerges, the gate progressively opens, allowing appearance residuals to refine rendering quality. This per-Gaussian design is particularly suitable for dynamic scenes with non-uniform motion difficulty, where different regions may become motion-active at different times.

\vspace{-0.5em}
\subsection{Continuous Motion Evolution}
\label{2}
This module aims to learn motion states that evolve continuously over time. 
Since real-world physics is highly complex, we adopt a simplified yet effective formulation by treating each Gaussian particle as rigid, with fixed scale over time. 
Under this assumption, we introduce a Motion Network to predict the instantaneous motion states of each Gaussian particle at any time $t$, including linear velocity, linear acceleration, angular velocity, and angular acceleration: \begin{equation}
\left\{ \big({v}_{t},\,{a}_{t}\big),\; \big({\omega}_{t},\,{\eta}_{t}\big) \right\}
= M_{\theta}({x}_t,\,t)
\label{MotionNet}
\end{equation}
\textbf{Bidirectional Trajectory Sampling.}
Built upon the motion trajectory learned in Section~\ref{sec:motion_aware_traj}, for each observed timestamp $t$, we sample two neighboring instants on both sides with a small interval $\Delta t$, $t_l = t-\Delta t,  t_r = t+\Delta t$. 
The Gaussian states at $t_l$ and $t_r$ are directly obtained from the learned trajectory representation, and their motion states are predicted by the Motion Network. 
We then propagate both states bidirectionally toward the observation time $t$ using Runge-Kutta $2^{nd}$-order(RK2) method, bidirectional propagation yields two Gaussian estimates at the same observation time $t$, which are both rendered and supervised by the image at $t$. 
\begin{equation}
\hat{\mathcal{G}}_{t\leftarrow t_l} = \mathrm{RK2}\!\left[\mathcal{G}_{t_l},\, t_l,\, +\Delta t,\, M_\theta\right]
\qquad
\hat{\mathcal{G}}_{t\leftarrow t_r} = \mathrm{RK2}\!\left[\mathcal{G}_{t_r},\, t_r,\, -\Delta t,\, M_\theta\right]
\label{eq:bidirectional_rk2}
\end{equation}
However, supervising this propagation only with rendering loss is inherently \textbf{ambiguous}, because rendered appearance is a consequence of motion evolution rather than its cause. As a result, image supervision alone cannot uniquely determine the underlying motion states. This ambiguity is particularly severe in our setting, where the motion state includes not only position, but also velocity and acceleration. Therefore, learning motion dynamics purely from appearance matching remains an under-constrained inverse problem. To further constrain the propagated motion states, we impose bidirectional kinematic continuity across neighboring timestamps. Specifically, for each observed timestamp $t$, the motion states predicted at $t_l=t-\Delta t$ and $t_r=t+\Delta t$ are required to remain consistent when propagated toward the same observation time through the kinematic chain from acceleration to velocity and from velocity to displacement. This gives the following continuity losses:
\vspace{-0.5em}
\begin{align}
\ell_{\text{cont1}} &= \frac{1}{|\mathcal{G}|}\sum_{i\in\mathcal{G}}
\left\|
{v}_{t_l}^{i}-{v}_{t_r}^{i}
+\Delta t\!\left({a}_{t_l}^{i}+{a}_{t_r}^{i}\right)
\right\|_1 \\
\ell_{\text{cont2}} &= \frac{1}{|\mathcal{G}|}\sum_{i\in\mathcal{G}}
\left\|
{x}_{t_l}^{i}-{x}_{t_r}^{i}
+\Delta t\!\left({v}_{t_l}^{i}+{v}_{t_r}^{i}\right)
\right\|_1
\label{eq:kinematic_cont}
\end{align}
where $\mathcal{G}$ denotes the set of Gaussians, and
${x}_{t_{l/r}}^{i}$, ${v}_{t_{l/r}}^{i}$, and ${a}_{t_{l/r}}^{i}$
are the displacement, velocity, and acceleration of the $i$-th Gaussian
at $t_l$ and $t_r$, respectively.
Equivalently, these losses penalize the mismatch of the two-sided predictions when they are propagated to the same observation time $t$, thereby regularizing temporally coherent motion evolution beyond rendering supervision alone.

\subsection{Training}
\label{3}
The training of our framework consists of two parts: learning reliable deformation at observed timestamps, and then learning physical motion states on top of these trajectories.
We first optimize the canonical Gaussians together with the deformation field, and then jointly optimize the Motion Network and the deformation field with both rendering supervision and kinematic continuity constraint.

We begin with a canonical warm-up at the initial timestamp, where the canonical Gaussian parameters are optimized using all observations at $t=0$.
After this initialization, we train the deformation field under a {gradual temporal expansion} strategy.
Instead of introducing all observed timestamps at once, we progressively enlarge the supervised temporal support by adding one time window after another. Deformation field is optimized only on the currently revealed observations:
\begin{equation}
    \{\mathcal{G}, F_{\text{defo}}\}
    \leftarrow
    (\ell_{1}+\ell_{\text{ssim}})
\end{equation}
This gradual schedule is naturally matched with our motion-guided appearance design.
As the supervised window expands, the per-Gaussian motion confidence $q_i$ is updated from the local deformation change between adjacent timestamps.
Consequently, different Gaussians in different motion regions can adaptively open or suppress the freedom of the canonical appearance parameters $({\sigma}_i,{c}_i)$ at different stages of optimization.
This prevents static or motion-inactive regions from reducing rendering loss through appearance shortcuts, while allowing genuinely motion-active regions to refine appearance once deformation becomes reliable.

After establishing reliable deformation trajectories, we train motion network with bidirectional sampling.
For each timestamp $t$, we sample $t_l=t-\Delta t$ and $t_r=t+\Delta t$, propagate their Gaussian states to $t$ using RK2 (Section~\ref{2}), and supervise the rendered results at $t$ with additional bidirectional kinematic continuity constraints.
The overall objective is:
\begin{equation}
    \{M_{\theta}, F_{\text{defo}}\}
    \leftarrow
    (\ell_{1}+\ell_{\text{ssim}})
    + \lambda_{\text{cont1}}\ell_{\text{cont1}}
    + \lambda_{\text{cont2}}\ell_{\text{cont2}}
\end{equation}

\section{Experiments}
\label{sec:experiments}
We evaluate our method on two tasks: \textbf{Geometrically Faithful Interpolation} and \textbf{Continuous Motion extrapolation}. 
Experiments are conducted on four dynamic scene datasets, covering both real-world and synthetic scenarios with diverse motion patterns and physics-related dynamics.

\textbf{Datasets.}
We use four datasets for evaluation.
1\text{)} {NVIDIA Dynamic Scenes Dataset}~\cite{wang2021neural}. 
We select two real-world dynamic scenes, including a person skateboarding and a moving truck, to evaluate performance on real-world dynamic reconstruction.
2\text{)} {TRACE MultiPart Dataset}~\cite{li2025trace}. 
This synthetic dataset contains multi-part objects with piecewise motions and interpolation/extrapolation splits, which are used to evaluate motion continuity under articulated dynamics.
3\text{)} {FreeGave-GoPro Dataset}~\cite{li2025freegave}. 
This dataset contains real-world dynamic scenes captured by a multi-camera GoPro rig, including challenging object interactions such as cube collisions.
4\text{)} PhysInOne Dataset~\cite{zhou2026physinone} 
contains synthetic scenes with explicit physical effects. 
We select six scenes covering friction, gravity, wind force, buoyancy physical mechanisms to evaluate dynamic reconstruction and future prediction.

\textbf{Baselines:}
We compare against eight representative baselines, grouped by their dynamic modeling strategies.
1\text{)} \textbf{Spatial-temporal reconstruction methods.}
{DefGS}~\cite{yang2024deformable}, 
{TiNeuVox}~\cite{fang2022fast}, 
{4DGS}~\cite{wu20244d}, and 
{ModecGS}~\cite{kwak2025modec}
model dynamic scenes mainly through spatial-temporal representations, making them strong baselines for reconstruction and interpolation within the observed time span.
2\text{)} \textbf{Physics-oriented motion modeling.}
{NVFi}~\cite{li2023nvfi} learns velocity-based motion fields for future extrapolation, but is less focused on high-fidelity spatial-temporal reconstruction.
3\text{)} \textbf{Hybrid trajectory-guided physical motion learning.}
{FreeGave}~\cite{li2025freegave}, 
{TRACE}~\cite{li2025trace}, and 
{ODEGS}~\cite{wang2025ode}
combine trajectory or deformation representations with physical-motion modeling, making them the closest baselines to our setting.
Together, these methods cover reconstruction-oriented, prediction-oriented, and hybrid physics-aware dynamic scene modeling, enabling a comprehensive evaluation.

\begin{wraptable}{r}{0.57\textwidth}
\vspace{-0.7em}
\scriptsize
\caption{Geometrically Faithful Interpolation Evaluation on NVIDIA Dynamic and PhysInOne Datasets. The best metric is highlight in red, the second best in orange.}
\centering
\setlength{\tabcolsep}{1.85pt}
\renewcommand{\arraystretch}{1.05}
\begin{tabular}{@{}lccc@{\hspace{5pt}}cccc@{}}
\toprule
 &
\multicolumn{3}{c}{\textbf{NVIDIA Dynamic Scene}} &
\multicolumn{4}{c}{\textbf{PhysInOne}} \\
\cmidrule(lr){2-4}\cmidrule(lr){5-8}
\rowcolor{headergray}
Method
& PSNR$\uparrow$ & SSIM$\uparrow$ & LPIPS$\downarrow$
& PSNR$\uparrow$ & SSIM$\uparrow$ & LPIPS$\downarrow$ & CD (m)$\downarrow$ \\
\midrule
TiNeuVox~\cite{fang2022fast}       
& \second{30.304} & 0.868 & 0.216 
& \second{30.923} & \second{0.957} & 0.123 & 0.641 \\

4DGS~\cite{wu20244d}               
& 24.411 & 0.462 & 0.532 
& 28.940 & 0.941 & 0.141 & 0.544 \\

DefGS~\cite{yang2024deformable}    
& 26.662 & \second{0.893} & 0.127 
& 30.165 & 0.944 & 0.137 & 0.877 \\

FreeGave~\cite{li2025freegave}     
& 30.276 & 0.891 & 0.096 
& 30.850 & 0.952 & 0.124 & \second{0.496} \\

ModecGS~\cite{kwak2025modec}       
& 30.080 & \best{0.914} & \best{0.081} 
& 30.110 & 0.938 & \second{0.113} & 0.678 \\

\midrule
\textbf{Ours}              
& \best{31.825} & \best{0.914} & \second{0.089} 
& \best{32.631} & \best{0.971} & \best{0.090} & \best{0.046} \\
\bottomrule
\end{tabular}
\label{interpolation_tab}
\vspace{-1.0em}
\end{wraptable}
\textbf{Metrics:}
We evaluate both RGB rendering quality and dynamic geometric accuracy.
For rendering, we report \textbf{PSNR}, \textbf{SSIM}, and \textbf{LPIPS}.
For geometry, we report \textbf{Chamfer Distance (CD)} on dynamic point clouds.
Specifically, PhysInOne provides ground truth depth maps at each timestamp, from which we back project 3D point clouds and extract dynamic regions through static/dynamic separation.
We compute CD between the reconstructed and ground truth dynamic point clouds, allowing us to evaluate whether the dynamic scene geometry is reconstructed at the correct 3D locations.

\begin{figure*}[]
  \centering
  \includegraphics[width=5.6in]{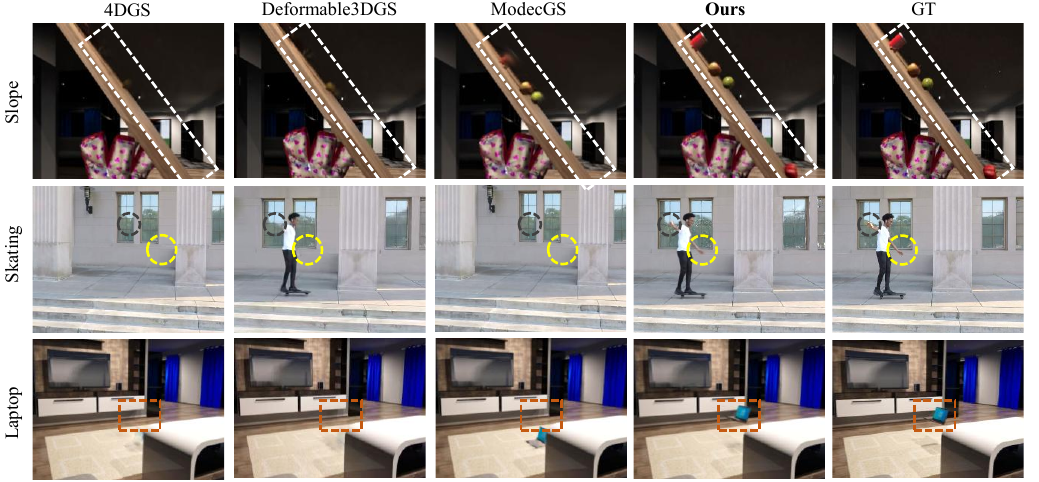}
  \caption{Qualitative Results of Geometrically Faithful Interpolation Task.}
  \label{fig:dynamics_factor_graph}
\end{figure*}
\subsection{Evaluation for Geometrically Faithful Interpolation}
\label{sec:kinematics}

We evaluate interpolation performance on NVIDIA Dynamic Scene and PhysInOne with representative baselines, including {TiNeuVox}, {4DGS}, {Deformable3DGS}(DefGS), {ModecGS}, and {FreeGave}. 
The time span of each sequence is normalized to one time unit.
Besides standard RGB metrics, we additionally report \textbf{Chamfer Distance (CD)} on PhysInOne to evaluate the 3D geometric accuracy of reconstructed dynamics. This metric is particularly important for dynamic reconstruction, as high rendering quality does not imply that the underlying dynamic geometry follows the correct motion.

\textbf{Results and Analysis.}
As shown in Table~\ref{interpolation_tab}, our method achieves the best interpolation performance in both RGB rendering metrics and dynamic geometric accuracy.
Notably, the CD metric is substantially improved over current methods, indicating that our method better recovers the spatial positions and geometry of dynamic regions rather than merely fitting visual appearance.
The key reason is that our framework uses local motion derived from the deformation field to guide the learning of the canonical representation, which reduces common appearance shortcuts in dynamic Gaussian reconstruction.
In particular, opacity $\sigma$ can suppress poorly tracked dynamic Gaussians and make moving regions disappear, while color $\mathbf{c}$  can compensate image-level errors without placing the dynamic geometry at correct 3D positions.
These shortcuts may still produce plausible renderings, but they lead to incorrect dynamic trajectories and are exposed by the Chamfer Distance on dynamic point clouds.
By explicitly coupling deformation with motion states, our method better constrains the spatial evolution of dynamic Gaussians, leading to both higher rendering quality and more faithful geometry.
This advantage is also reflected in the qualitative results in Figure~\ref{fig:dynamics_factor_graph}.
In the  {Laptop} and {Skating} scene, competing methods tend to lose dynamic parts due to opacity driven disappearance, whereas our method preserves the moving geometry.
In the Slope scene, other methods can partially fit the appearance through color compensation but fail to recover the correct dynamic structure or fine grained motion, while our method reconstructs more coherent trajectories and sharper geometry.

\begin{table}[]
\caption{Quantitative Results for Continuous Motion Extrapolation on All Four Datasets. The best metric is highlight in red, while the second best highlight in orange.}
 \vspace{0.5em}
\centering
\footnotesize
\setlength{\tabcolsep}{1.0pt}
\renewcommand{\arraystretch}{1.15}
\begin{tabular}{@{}lccc@{\hspace{3.5pt}}ccc@{\hspace{3.5pt}}ccc@{\hspace{3.5pt}}cccc@{}}
\toprule
&
\multicolumn{3}{c}{\textbf{NVIDIA Dynamic Scene}} &
\multicolumn{3}{c}{\textbf{Dynamic Multipart}} &
\multicolumn{3}{c}{\textbf{FreeGave-GoPro}} &
\multicolumn{4}{c}{\textbf{PhysInOne}} \\
\cmidrule(l{1pt}r{1pt}){2-4}
\cmidrule(l{1pt}r{1pt}){5-7}
\cmidrule(l{1pt}r{1pt}){8-10}
\cmidrule(l{1pt}r{1pt}){11-14}
\rowcolor{headergray}
Method
& PSNR$\uparrow$ & SSIM$\uparrow$ & LPIPS$\downarrow$
& PSNR$\uparrow$ & SSIM$\uparrow$ & LPIPS$\downarrow$
& PSNR$\uparrow$ & SSIM$\uparrow$ & LPIPS$\downarrow$
& PSNR$\uparrow$ & SSIM$\uparrow$ & LPIPS$\downarrow$ & CD (m)$\downarrow$ \\
\midrule
TiNeuVox~\cite{fang2022fast}
& 24.556 & 0.863 & 0.215
& 20.804 & 0.923 & 0.090
& 20.600 & 0.760 & 0.292
& 19.264 & 0.884 & 0.254 & 0.925 \\

NVFi~\cite{li2023nvfi}
& 28.462 & 0.876 & 0.214
& 25.235 & 0.955 & 0.046
& 22.747 & 0.756 & 0.393
& 25.571 & 0.902 & 0.131 & 0.610 \\

DefGS~\cite{yang2024deformable}
& 24.240 & 0.895 & 0.140
& 20.664 & 0.930 & 0.067
& 21.344 & 0.858 & 0.170
& 20.384 & 0.867 & 0.213 & 1.029 \\

4DGS~\cite{wu20244d}
& 22.510 & 0.703 & 0.408
& 20.564 & 0.935 & 0.067
& 22.035 & 0.871 & 0.198
& 20.687 & 0.876 & 0.283 & 0.947 \\


ODEGS~\cite{wang2025ode}
& 25.296 & 0.916 & 0.117
& 33.371 & 0.982 & 0.009
& 25.561 & 0.899 & 0.124
& 24.023 & 0.916 & 0.121 & 0.854 \\

FreeGave~\cite{li2025freegave}
& 29.005 & \second{0.933} & \second{0.072}
& {33.183} & 0.980 & 0.009
& \second{28.094} & \second{0.914} & \second{0.112}
& \second{27.801} & 0.923 & 0.125 & 0.718 \\

TRACE~\cite{li2025trace}
& \second{29.341} & \second{0.933} & 0.074
& \second{33.481} & \second{0.990} & \second{0.007}
& 27.871 & 0.903 & 0.126
& 27.038 & \second{0.934} & \second{0.096} & \second{0.491} \\

\midrule
\textbf{Ours}
& \best{30.265} & \best{0.939} & \best{0.065}
& \best{35.754} & \best{0.991} & \best{0.007}
& \best{28.135} & \best{0.916} & \best{0.110}
& \best{29.591} & \best{0.949} & \best{0.091} & \best{0.073} \\
\bottomrule
\end{tabular}
\label{tab:future_extrapolation_all}
\vspace{-1.0em}
\end{table}

\subsection{Evaluation for Continuous Motion Extrapolation}
For the Continuous Motion extrapolation task, we focus on baselines that possess explicit predictive capability. Specifically, we compare against {FreeGave}, {TRACE} and {ODEGS}, which model the motion state of the scene, as well as {TiNeuVox}, {NVFi} , {4DGS} and {Deformable3DGS(DefGS), which parameterize space-time dynamics with NeRF-based and Gaussian-based representations, respectively, and thus can perform extrapolation to some extent. For the experimental protocol, we follow the prediction settings of {FreeGave} and {TRACE}. Specifically, we normalize the observation time to unit one. We observe frames in $[0,0.75]$ and predict future frames in $[0.75,1]$. 

\textbf{Results and Analysis.}
As shown in Table~\ref{tab:future_extrapolation_all}, our method achieves the best performance on future extrapolation across all datasets.
This advantage mainly comes from our continuity loss, which explicitly constrains the kinematic relationship among displacement, velocity, and acceleration.
Rendering supervision alone can fit observed frames, but it does not uniquely determine the underlying motion states, making the learned velocity and acceleration under-constrained for future prediction.
As a result, prior predictive baselines may recover reasonable appearances at observed times but still produce unstable or physically implausible extrapolated motion.
\begin{figure}[h]
  \vspace{-0.6em}
  \centering
  \includegraphics[width=5.6in]{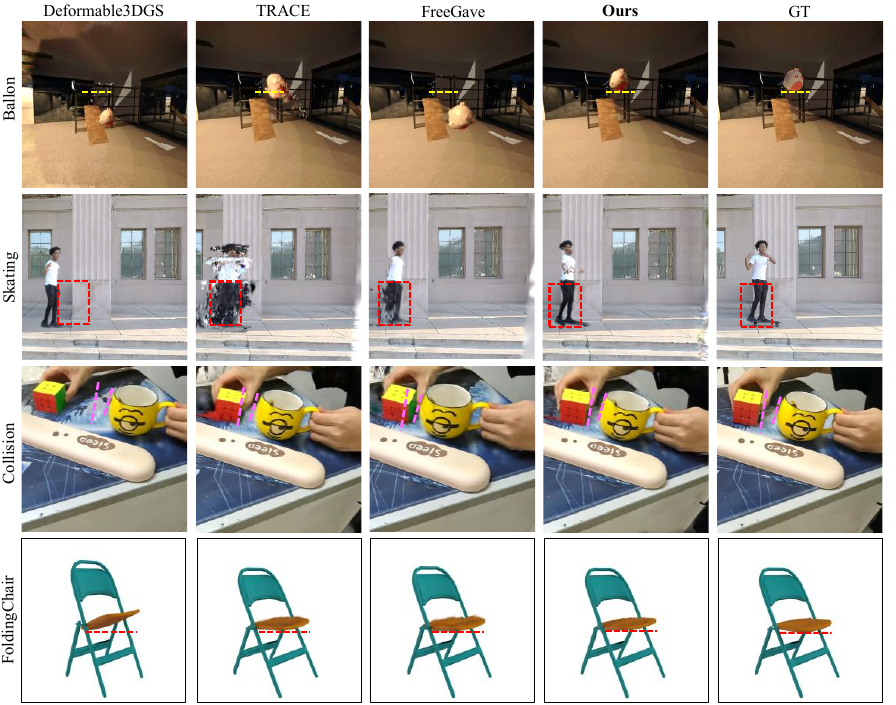}
  \caption{Qualitative Results of Continuous Motion Extrapolation on Four Datasets.}
  \label{fig:comparison}
  \vspace{-0.7em}
\end{figure}
By enforcing consistency from acceleration to velocity and from velocity to displacement, our method learns a more coherent motion evolution, leading to kinematically continuous and physically plausible predictions.
The qualitative results in Figure~\ref{fig:comparison} clearly reflect this effect.
In the {Balloon} scene, FreeGave predicts an incorrect motion direction, while our method correctly captures the upward buoyancy driven trend.
In the {Skating} scene, FreeGave and TRACE produce discontinuous future motion, whereas our method maintains a smoother and more coherent trajectory.
These results show that explicit kinematic continuity is essential for reliable future extrapolation.

\vspace{-0.5em} 
\subsection{Ablation Study}
\label{sec:ablation}
To validate the effectiveness of each component in our framework, we conduct the following ablation studies.  All ablation experiments are conducted on the PhysInOne Dataset.

\textbf{(1) Removing Motion Guidance:} 
remove the motion confidence $q_i$ used to guide the training of canonical appearance attributes. 
Instead, the canonical Gaussian parameters and the deformation field are optimized together without motion guidance. 

\textbf{(2) Removing Continuity Loss $\ell_{\text{cont1}}$ and $\ell_{\text{cont2}}$:} 
when jointly optimizing $f_{\text{defo}}$ and  $M_{\theta}$, we remove the continuity losses and train the model only with rendering supervision. 

\textbf{(3) Removing Motion Network :}
remove $M_{\theta}$ during evaluation and directly use $f_{\text{defo}}$.

\textbf{(4) Different Choices of Time Difference $\Delta t$:} 
 Given the time interval between two consecutive
frames in the training set as $\delta t$, compare different training intervals $\Delta t = \{\delta t, 2\delta t, 3\delta t\}$, where $\delta t$ denotes the time interval between two adjacent frames.

\textbf{Results and Analysis.}
From Table~\ref{tab:ablation_interp}, we observe that each key design consistently contributes to the final performance.
\textbf{(1)} Removing motion guidance $q_i$ leads to  obvious degradation, especially in geometry-related accuracy. This supports our motivation that appearance optimization without motion awareness can fall into shortcut solutions. Since the learned trajectory further serves as the basis for estimating velocity and acceleration, this degradation also affects future prediction.
\begin{wraptable}{r}{0.43\columnwidth}
\vspace{-0.8em}
\caption{Quantitative Results of Ablation Studies. The best metric is highlight in red, while the second best highlight in orange.}
\centering
\scriptsize
\setlength{\tabcolsep}{1.05pt}
\renewcommand{\arraystretch}{1.22}
\resizebox{\linewidth}{!}{
\begin{tabular}{@{}lcccc@{\hspace{4pt}}ccc@{\hspace{4pt}}ccc@{}}
\toprule
&
\multicolumn{4}{c}{} &
\multicolumn{3}{c}{\textbf{Interp.}} &
\multicolumn{3}{c}{\textbf{Extrap.}} \\
\cmidrule(lr){6-8}\cmidrule(lr){9-11}
\rowcolor{headergray}
& $q_i$ 
& $\ell_{\text{cont1}}$ 
& $\ell_{\text{cont2}}$
& $M_{\theta}$
& PSNR$\uparrow$ & SSIM$\uparrow$ & CD (m)$\downarrow$
& PSNR$\uparrow$ & SSIM$\uparrow$ & CD (m)$\downarrow$ \\
\midrule
(1) & \xmark & \cmark & \cmark & \cmark
& 30.164 & 0.909 & 0.540
& 29.016 & 0.924 & 0.358 \\

(2) & \cmark & \xmark & \cmark & \cmark
& 31.927 & 0.965 & 0.066
& 28.945 & 0.921 & 0.079 \\

(2) & \cmark & \cmark & \xmark & \cmark
& \second{32.628} & 0.934 & 0.047
& 27.468 & 0.914 & 0.151 \\

(3) & \cmark & \xmark & \xmark & \xmark
& 32.521 & \second{0.970} & \second{0.046}
& 23.297 & 0.901 & 0.547 \\

\midrule
$1\delta t$ & \cmark & \cmark & \cmark & \cmark
& 32.498 & 0.956 & 0.059
& 29.411 & \second{0.947} & \second{0.075} \\

$2\delta t$ & \cmark & \cmark & \cmark & \cmark
& \best{32.631} & \best{0.971} & \best{0.046}
& \best{29.591} & \best{0.949} & \best{0.073} \\

$3\delta t$ & \cmark & \cmark & \cmark & \cmark
& {32.597} & {0.967} & 0.048
& \second{29.467} & 0.920 & 0.081 \\

\bottomrule
\end{tabular}
}
\label{tab:ablation_interp}
\vspace{-1.4em} 
\end{wraptable}
\textbf{(2)} Removing either continuity loss weakens extrapolation performance. This confirms that rendering supervision alone is insufficient to constrain the learned motion states: even if individual frames can be fitted, the transitions among position, velocity, and acceleration may become temporally inconsistent.
\textbf{(3)} Without $M_{\theta}$, the model does not learn physical motion states like velocity and acceleration, and thus loses its ability to predict physically plausible future dynamics.
\textbf{(4)} Different choices of $\Delta t$ produce relatively stable results, indicating that the proposed motion evolution formulation is not sensitive to moderate changes in the temporal interval.
Overall, these ablations show that motion-aware trajectory representation improves geometrically faithful interpolation, while continuous motion evolution strengthens future extrapolation. The two modules are complementary: more reliable trajectories provide better supervision for motion-state learning, and learned continuous motion states further improve dynamic reconstruction.
\vspace{-0.5em}

\section{Conclusion}
We introduced \textbf{PhyMo}, a novel framework that aims to recover accurate and continuous evolving dynamics rather than merely fitting visual appearance.
PhyMo combines a Motion-Aware Trajectory Representation to reduce appearance shortcuts with a Continuous Motion Evolution module to enforce temporal kinematic consistency.
Experiments on four dynamic scene benchmarks show that our method achieves strong interpolation and extrapolation performance, producing more accurate dynamic geometry and more continuous, physically plausible future predictions.
\begin{ack}
This work is supported in part by the Natural Science Foundation of China
(42301520), the Research Grants Council of Hong Kong (15212925, 25206524). The authors declare no competing interests.
\end{ack}
\bibliography{nips2026}
\bibliographystyle{unsrtnat}

\clearpage
\newpage
\appendix
\onecolumn
\section{Appendix}

The appendix includes:
\begin{itemize}[topsep=2pt, itemsep=1pt, parsep=0pt, partopsep=0pt]
    \item More Details of Motion-Aware Trajectory Representation (Appendix~\ref{app:motion_aware_tra})
    \item Implementation Details of Continuous Motion Evolution Module (Appendix~\ref{app:motion module})
    \item Implementation Details of Training (Appendix~\ref{app:training_schedule}) 
    \item Additional Ablation Results (Appendix~\ref{app:ablation})
    \item Additional Qualitative Results (Appendix~\ref{app:qualitative})
    \item Additional Quantitative Results (Appendix~\ref{app:quantitative})
    \item Details of All Datasets (Appendix~\ref{app:datasets})
    \item Training and Evaluation Resources (Appendix~\ref{app:resource})
    \item Limitation of Our Model (Appendix~\ref{limitation})
\end{itemize}
In addition, we provide a \textbf{demo video} for visualization,
appended in the supplementary materials.
\subsection{More Details of Motion-Aware Trajectory Representation}
\label{app:motion_aware_tra}

\paragraph{Implementation details of motion confidence.}
The main paper defines the motion-guided appearance parameterization in Equation~\eqref{eq:motion_aware_app} and the per-Gaussian motion confidence in Equation~\eqref{eq:motion_confidence}. 
Here we provide the exact implementation used in our experiments. 
When the motion-aware appearance branch is activated, we first store frozen anchors of the canonical appearance parameters, including opacity and SH coefficients. 
Specifically, let $\sigma_i^{b}$, ${c}_{i,\mathrm{dc}}^{b}$, and ${c}_{i,\mathrm{rest}}^{b}$ denote the anchored opacity, DC SH term, and higher-order SH terms of the $i$-th Gaussian, respectively. 
During training, the appearance update is released from these anchors according to the motion confidence $q_i$.
The motion confidence is computed from the local deformation change between two adjacent timestamps, consistent with Equation~\eqref{eq:motion_confidence}. 
We instantiate the monotonic function $\psi(\cdot)$ as a sigmoid gate:
\begin{equation}
q_i
=
\mathrm{sg}\!\left[
q_{\min} + (1-q_{\min})\,
\mathrm{sigmoid}\!\left(
\beta\left(
\left\|
\delta {x}_i(t_{k+1})-\delta {x}_i(t_k)
\right\|_2
-\tau
\right)
\right)
\right]
\label{eq:q_impl_appendix}
\end{equation}
where $\tau$ is the gate threshold, $\beta$ controls the sharpness of the transition, $q_{\min}$ is the minimum gate value, and $\mathrm{sg}(\cdot)$ denotes stop-gradient. 
In all reported experiments, we use $\tau=0.25$, $\beta=50$, and $q_{\min}=0.0001$. 
With this setting, the gate stays nearly closed when the local deformation change is weak, and opens rapidly once clear local motion activation emerges.
\begin{figure*}[htbp]
  \centering
  \includegraphics[width=5.5in]{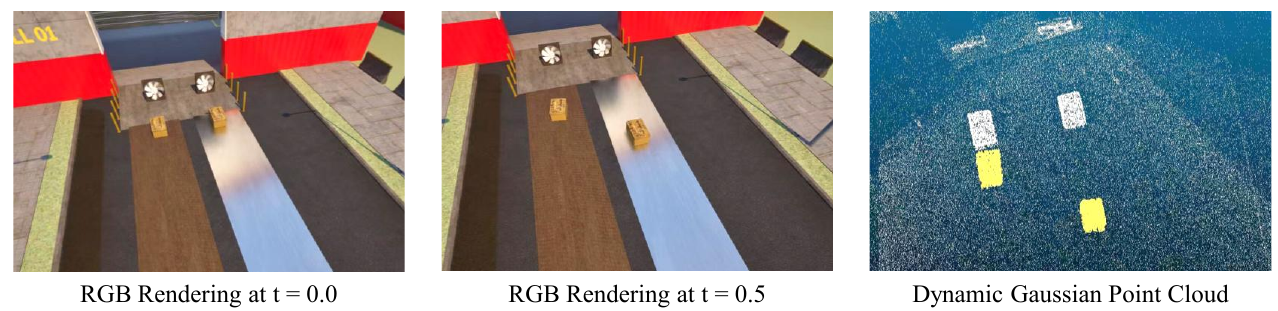}
  \caption{
RGB Rendering and the Evolving Geometry of Sandwich Scene.
}
  \label{fig:sandwich}
  \vspace{-0.7em}
\end{figure*}
After computing $q_i$, the anchored appearance parameters are released by a residual form. 
For the SH coefficients, we separately gate the DC term and the remaining higher-order terms:
\begin{align}
{c}_{i,\mathrm{dc}} &=
{c}_{i,\mathrm{dc}}^{b}
+
s_{\mathrm{dc}}\,q_i
\big(
{c}_{i,\mathrm{dc}}-{c}_{i,\mathrm{dc}}^{b}
\big) \\
{c}_{i,\mathrm{rest}} &=
{c}_{i,\mathrm{rest}}^{b}
+
q_i
\big(
{c}_{i,\mathrm{rest}}-{c}_{i,\mathrm{rest}}^{b}
\big)
\end{align}
where $s_{\mathrm{dc}}$ is an optional scaling factor for the DC term, and we set $s_{\mathrm{dc}}=0.95$. 
The opacity branch follows the same anchor-plus-gated-residual form as Equation~\eqref{eq:motion_aware_app}. 
Importantly, the gate is detached from back-propagation, so that the appearance branch does not directly perturb the motion confidence.

\textbf{Effectiveness validated by training dynamics.}
We further verify the effectiveness of motion confidence by visualizing the training dynamics on the {Sandwich} scene, as shown in Figure~\ref{fig:sandwich}.
\begin{figure*}[htbp]
  \centering
  \includegraphics[width=5.6in]{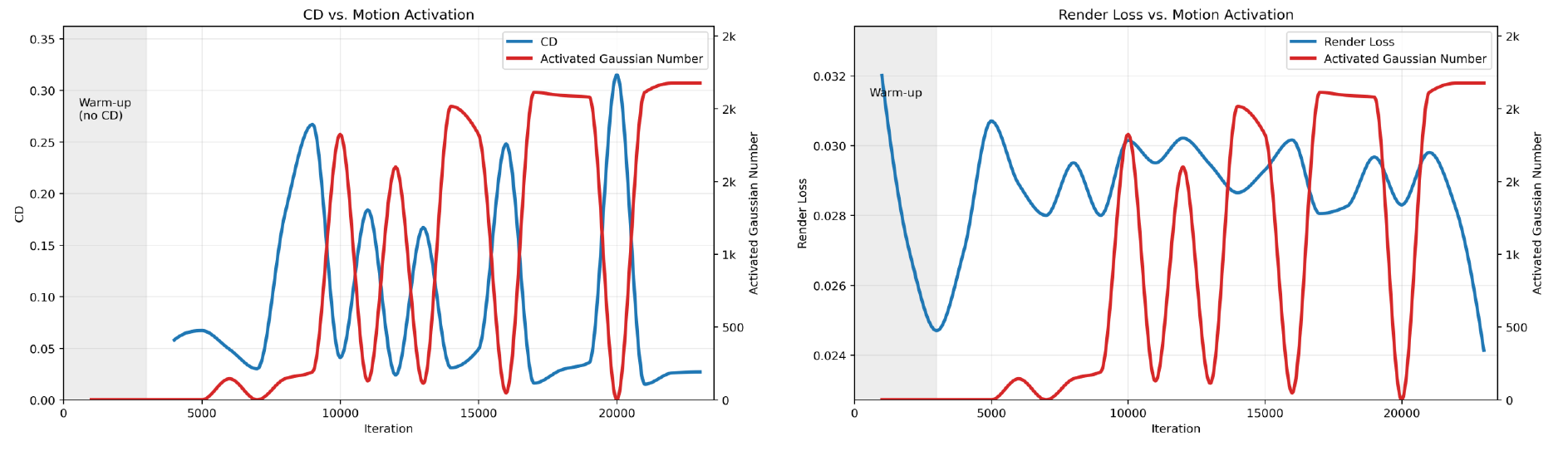}
  \caption{
Training Dynamics of Motion-Aware Trajectory Representation on the {Sandwich} Scene.
}
  \label{fig:sandwich_training_dynamics}
  \vspace{-1.0em}
\end{figure*}
In this experiment, we record the rendering loss, the Chamfer Distance (CD) between reconstructed and ground-truth dynamic point clouds, and the number of activated Gaussians during training.
The first 3,000 iterations correspond to the canonical Gaussian warm-up stage.

To summarize the per-Gaussian motion confidence during training, we report the activated Gaussian number, defined as $N_{\mathrm{act}}=\sum_{i\in\mathcal{G}}\mathbf{1}(q_i>\eta)$ with $\eta=0.5$.
Since $q_i$ is produced by the sigmoid gate in Equation~\eqref{eq:q_impl_appendix}, this threshold corresponds to the midpoint of the gate response; namely, a Gaussian is counted as activated once its local deformation change exceeds the gate threshold $\tau$.
We record $N_{\mathrm{act}}$ together with CD and rendering loss along the training iterations.
This is particularly informative under our gradual temporal expansion : as new temporal windows are progressively introduced, the evolution of CD, rendering loss, and activated Gaussian number reflects how motion becomes reliable at different timestamps and different scene regions.
\begin{table*}[htbp]
\centering
\vspace{-0.58em}
\caption{Ablation Study of Motion-Guided Appearance on PhysInOne.}
\vspace{-0.38em}
\label{tab:motion_confidence_ablation}
\scriptsize
\setlength{\tabcolsep}{4.8pt}
\renewcommand{\arraystretch}{1.08}
\resizebox{0.82\textwidth}{!}{
\begin{tabular}{@{}lcccccc@{}}
\toprule
&
\multicolumn{3}{c}{\textbf{Design}} &
\multicolumn{3}{c}{\textbf{Interp.}} \\
\cmidrule(lr){2-4}\cmidrule(lr){5-7}
\rowcolor{headergray}
Variant
& $q_i$
& Per-G.
& Local Motion
& PSNR$\uparrow$
& SSIM$\uparrow$
& CD (m)$\downarrow$ \\
\midrule

w/o $q_i$
& \xmark & \xmark & \xmark
& 30.164 & 0.909 & 0.540 \\

Freeze-then-release
& \xmark & \xmark & \xmark
& 31.285 & 0.902 & 0.267 \\

Disp.-based $q_i$
& \cmark & \cmark & \xmark
& 30.874 & 0.924 & 0.351 \\

Local-motion $q_i$
& \cmark & \cmark & \cmark
& 32.631 & 0.971 & 0.046 \\

\bottomrule
\end{tabular}
}
\vspace{-0.58em}
\end{table*}
As shown in Figure~\ref{fig:sandwich_training_dynamics}, after the warm-up stage, the activated Gaussian number remains small when reliable motion has not yet been established, while CD stays relatively high.
As training proceeds, the activated Gaussian number increases at several stages, and these increases consistently coincide with decreases in CD.
This indicates that the motion confidence gate opens when local deformation becomes reliable, rather than releasing appearance freedom from the beginning.
The rendering loss then decreases clearly after more Gaussians become activated, showing that appearance refinement is applied after the model has recovered  accurate  trajectories.
The repeated activation pattern also reflects the non-uniform difficulty of different motion regions, where different parts become motion-active at different time.

\textbf{More Ablation.} We further conduct ablation experiments on PhysInOne to analyze the design of motion-guided appearance.
We compare four variants:
(1) \textbf{w/o $q_i$}, which jointly optimizes canonical appearance and deformation without motion confidence;
(2) \textbf{Freeze-release}, which globally freezes canonical appearance for the first $10{,}000$ iterations and then jointly optimizes appearance and deformation for another $1{,}0000$ iterations;
(3) \textbf{Disp.-based $q_i$}, which computes motion confidence from the cumulative deformation magnitude, $q_i=\psi(\|\delta x_i(t)\|_2)$;
and (4) \textbf{Local-motion $q_i$}, our final design that computes motion confidence from local deformation change between adjacent timestamps.

As shown in Table~\ref{tab:motion_confidence_ablation}, removing motion confidence clearly degrades both rendering quality and dynamic geometric accuracy, indicating that fully free canonical appearance can  act as a shortcut during deformation learning.
The freeze-release strategy improves stability by delaying appearance optimization, but it applies the same schedule to all Gaussians and therefore cannot adapt to spatially varying motion difficulty.
In contrast, our per-Gaussian confidence releases appearance freedom according to the actual local motion response of each Gaussian.
Compared with the cumulative-displacement variant, local deformation change is a more accurate indicator of current motion activity: a Gaussian may have moved far from the canonical state simply due to accumulated displacement, while adjacent deformation change directly reflects whether it is actively participating in the current motion.
These results show that the proposed motion confidence is important for suppressing appearance shortcuts and improving the accuracy of recovered dynamic geometry.

\subsection{Implementation Details of Continuous Motion Evolution Module}
\label{app:motion module}
To estimate per-particle motion states, we employ a Motion Network that predicts translational and rotational motion parameters for each Gaussian particle at timestamp $t$:
\begin{equation}
\left\{ \big({v}_{t},\,{a}_{t}\big),\; \big({\omega}_{t},\,{\eta}_{t}\big) \right\}
= M_{\theta}({x}_t,\,t)
\label{eq:motion_net_app}
\end{equation}
where $(v_t,a_t)\in\mathbb{R}^{6}$ denote the translational velocity and acceleration, and $(\omega_t,\eta_t)\in\mathbb{R}^{6}$ denote the angular velocity and angular acceleration, respectively.
The motion network is implemented by an $8{\times}256$ MLP with ReLU activation, consisting of 8 fully-connected layers and 256 hidden units per layer.
An 8-degree positional embedding is applied to the input Gaussian position $x_t$ before feeding it into the MLP, enabling the network to model spatially varying motion states.
An 8-degree positional embedding is applied to the input 3D position before feeding into the MLP, enabling the network
to model spatially-varying motion states of the dynamic scene.
\begin{algorithm}[t]
\caption{RK2-based Bidirectional Motion Propagation}
\label{alg:rk2_motion_integration}
\small
\begin{algorithmic}[1]
\Require Canonical Gaussians 
$\mathcal{G}_0=\{\cdots(x_0,r_0,s_0,\sigma,c)\cdots\}$, target timestamp $t$, step size $\Delta t$, deformation field $f_{\mathrm{defo}}$, motion network $M_\theta$
\Ensure Integrated Gaussians 
$\mathcal{G}_t=\{\cdots(x_t,r_t,s_t,\sigma,c)\cdots\}$

\State $t_l \gets t-\Delta t,\quad t_r \gets t+\Delta t$

\State $(\delta x_{t_l},\delta r_{t_l},\delta s_{t_l})
\gets f_{\mathrm{defo}}(x_0,e(t_l))$
\State $(\delta x_{t_r},\delta r_{t_r},\delta s_{t_r})
\gets f_{\mathrm{defo}}(x_0,e(t_r))$

\State $x_{t_l}\gets x_0+\delta x_{t_l},\quad
x_{t_r}\gets x_0+\delta x_{t_r}$
\State $r_{t_l}\gets r_0\circ\delta r_{t_l},\quad
r_{t_r}\gets r_0\circ\delta r_{t_r}$

\State $\left\{(v_{t_l},a_{t_l}),(\omega_{t_l},\eta_{t_l})\right\}
\gets M_\theta(x_{t_l},t_l)$
\State $\left\{(v_{t_r},a_{t_r}),(\omega_{t_r},\eta_{t_r})\right\}
\gets M_\theta(x_{t_r},t_r)$

\State $v_{\mathrm{mid}}^{l}\gets v_{t_l}+\frac{\Delta t}{2}a_{t_l},\quad
v_{\mathrm{mid}}^{r}\gets v_{t_r}-\frac{\Delta t}{2}a_{t_r}$

\State $\omega_{\mathrm{mid}}^{l}\gets \omega_{t_l}+\frac{\Delta t}{2}\eta_{t_l},\quad
\omega_{\mathrm{mid}}^{r}\gets \omega_{t_r}-\frac{\Delta t}{2}\eta_{t_r}$

\State $\hat{x}_{t}^{\,l}
\gets x_{t_l}+\Delta t\left(v_{\mathrm{mid}}^{l}
+\omega_{\mathrm{mid}}^{l}\times x_{t_l}\right)$

\State $\hat{x}_{t}^{\,r}
\gets x_{t_r}-\Delta t\left(v_{\mathrm{mid}}^{r}
+\omega_{\mathrm{mid}}^{r}\times x_{t_r}\right)$

\State $\Delta r^{l}
\gets \exp_q\!\left(\frac{1}{2}\omega_{\mathrm{mid}}^{l}\Delta t\right)$
\State $\Delta r^{r}
\gets \exp_q\!\left(-\frac{1}{2}\omega_{\mathrm{mid}}^{r}\Delta t\right)$

\State $\hat{r}_{t}^{\,l}
\gets \operatorname{normalize}\!\left(\Delta r^{l}\circ r_{t_l}\right)$
\State $\hat{r}_{t}^{\,r}
\gets \operatorname{normalize}\!\left(\Delta r^{r}\circ r_{t_r}\right)$

\State $x_t\gets \frac{1}{2}\left(\hat{x}_{t}^{\,l}+\hat{x}_{t}^{\,r}\right)$
\State $r_t\gets \operatorname{SLERP}\!\left(\hat{r}_{t}^{\,l},\hat{r}_{t}^{\,r};\,0.5\right)$
\State $s_t\gets s_0$

\State \Return $\mathcal{G}_t=\{\cdots(x_t,r_t,s_t,\sigma,c)\cdots\}$
\end{algorithmic}
\end{algorithm}

Algorithm \ref{alg:rk2_motion_integration} follows the second-order Runge-Kutta method, also known as the explicit midpoint rule.
For a motion state governed by a time-varying velocity, RK2 first estimates the velocity at the middle of the interval and then uses this midpoint velocity to propagate the state over the full step.
In our case, the translational midpoint velocity is computed as $v_{\mathrm{mid}}=v+\frac{\Delta t}{2}a$, while the angular midpoint velocity is computed as $\omega_{\mathrm{mid}}=\omega+\frac{\Delta t}{2}\eta$.
The Gaussian position and rotation are then propagated using these midpoint motion states, which provides a second-order approximation for local motion evolution.
We perform this propagation bidirectionally.
Starting from the left neighbor $t_l=t-\Delta t$, the Gaussian state is propagated forward to the target timestamp $t$.
Starting from the right neighbor $t_r=t+\Delta t$, the Gaussian state is propagated backward to $t$ with a negative time step.
The two propagated positions are averaged, while the two propagated rotations are fused by spherical linear interpolation, denoted as $\operatorname{SLERP}(\cdot)$.
Here, $\operatorname{SLERP}(r_1,r_2;\alpha)$ interpolates two unit quaternions along the shortest path on the rotation sphere, and we set $\alpha=0.5$ to take the midpoint rotation.
Moreover, $\exp(\cdot)$ denotes the $\mathrm{SO}(3)$ exponential map, and $\operatorname{normalize}(\cdot)$ indicates quaternion normalization to enforce unit norm.
The resulting propagated state is used for rendering supervision, while the two directional motion states are further constrained by the kinematic continuity losses.

\subsection{Implementation Details of Training Schedule}
\label{app:training_schedule}

We provide the detailed training schedule used in our experiments.
The training starts with a canonical Gaussian warm-up for $3{,}000$ iterations, where only the observations at the initial timestamp $t=0$ are used to initialize the canonical geometry and appearance.

After warm-up, we train the deformation field with gradual temporal expansion.
Specifically, we divide the observed time span into $N=10$ temporal windows.
Training begins from the first window, and then progressively includes one additional window until all observed timestamps are involved.
This deformation learning stage takes $20{,}000$ iterations in total.
During this process, the motion confidence $q_i$ is updated according to the local deformation change within the currently revealed temporal support, so the freedom of canonical appearance is released adaptively.

Finally, we jointly train the deformation field and the Motion Network for another $20{,}000$ iterations.
We use bidirectional neighboring timestamps and RK2-based propagation to obtain the Gaussian states at the supervised timestamp.
The rendering loss is optimized together with the bidirectional kinematic continuity losses.
In all experiments, we set the loss weights as $\lambda_{\mathrm{cont1}}=\lambda_{\mathrm{cont2}}=0.3$.
\subsection{Additional Ablation Results}
\label{app:ablation}
\begin{figure*}[h!]
  \centering
  \includegraphics[width=4.7in]{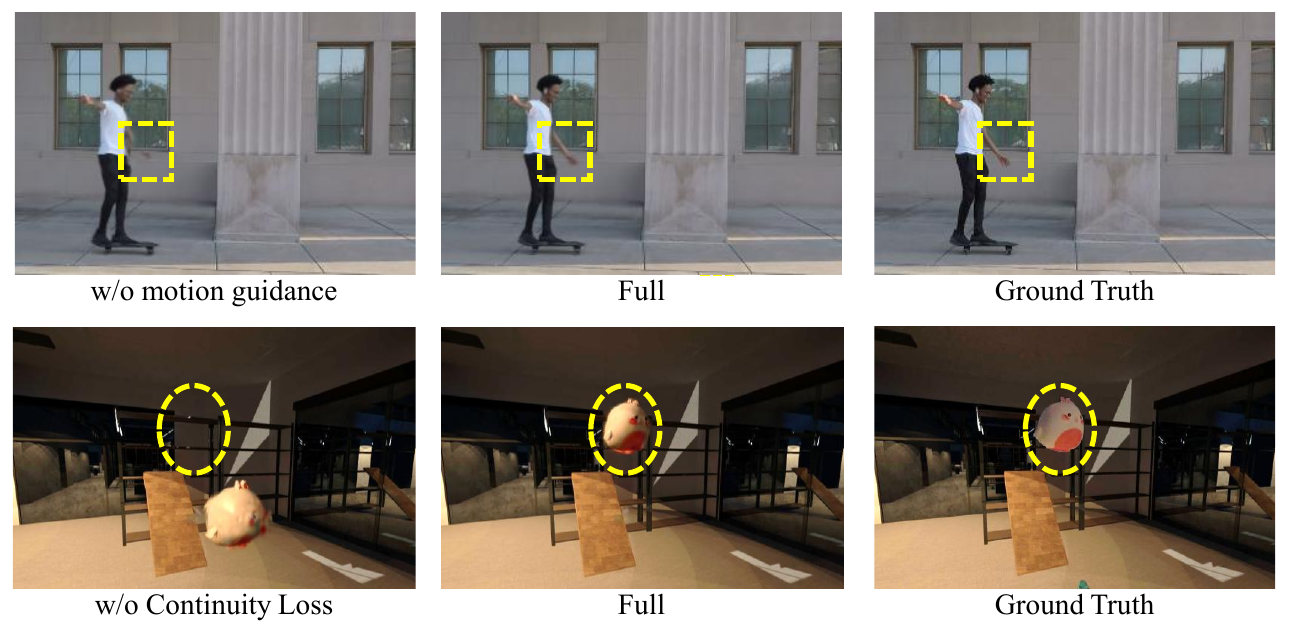}
  \caption{Qualitative  Results of Ablation about Motion Guidance and Continuity Loss.}
  \label{fig:ablation}
  \vspace{-0.7em}
\end{figure*}
Here, we further provide qualitative comparisons for a more intuitive understanding.
As shown in Figure~\ref{fig:ablation}, on the {Skating} scene for novel-time interpolation, removing motion guidance causes the moving arm to disappear, indicating that the model tends to reduce rendering error through appearance shortcuts instead of recovering the correct motion.
On the {Ballon} scene for future-frame extrapolation, removing the continuity loss leads to an incorrect moving direction of the balloon and breaks the temporal consistency of the predicted motion.
These qualitative results further demonstrate the importance of motion-guided appearance and kinematic continuity for faithful dynamic reconstruction and physically plausible prediction.
\begin{table}[h]
\centering
\scriptsize
\caption{Interpolation Evaluation on Dynamic MultiPart and FreeGave-GoPro Dataset.}
\vspace{0.3em}
\setlength{\tabcolsep}{5pt}
\renewcommand{\arraystretch}{1.3}
\begin{tabular}{l|ccc|ccc}
\toprule
 &
\multicolumn{3}{c|}{\textbf{Dynamic MultiPart}} &
\multicolumn{3}{c}{\textbf{FreeGave-GoPro}} \\
\cmidrule(lr){2-4}\cmidrule(lr){5-7}
& PSNR$\uparrow$ & SSIM$\uparrow$ & LPIPS$\downarrow$
& PSNR$\uparrow$ & SSIM$\uparrow$ & LPIPS$\downarrow$ \\
\midrule
TiNeuVox~\cite{fang2022fast} & 29.951 & 0.968 & 0.051 & 19.123 & 0.740 & 0.328 \\
4DGS~\cite{wu20244d}         & 37.025 & 0.992 & 0.016 & 28.940 & 0.941 & 0.141 \\
DefGS~\cite{yang2024deformable}       & 34.459 & 0.983 & 0.014 & 28.465 & 0.924 & 0.097 \\
FreeGave~\cite{li2025freegave} & 35.776 & 0.986 & 0.016 & 28.457 & 0.925 & 0.091 \\
ModecGS~\cite{kwak2025modec}       & 35.768 & 0.991 & 0.011 & 27.946 & 0.928 & 0.083 \\
\midrule
Ours                      & \textbf{37.625} & \textbf{0.994} & \textbf{0.009} & \textbf{29.231} & \textbf{0.951} & \textbf{0.079} \\
\bottomrule
\end{tabular}
\label{add_interpolation_tab}
\end{table}
\vspace{-0.5em}
\label{app:deformation}







\subsection{Additional Qualitative Results}
\label{app:qualitative}
Figure~\ref{Bus}, Figure~\ref{Freefall}, Figure~\ref{Writrest}, Figure~\ref{laptop} and Figure~\ref{skating} provides additional qualitative comparisons for both Geometrically Faithful Interpolation and Continuous Motion Extrapolation.
For geometrically faithful interpolation, as illustrated in the Laptop and Skating scenes, our method recovers more accurate intermediate geometry and motion than strong 4D reconstruction baselines.
For continuous motion extrapolation, as shown in the Bus and Freefall scenes, our method produces more accurate and temporally continuous future motion than FreeGave and TRACE, which tend to accumulate motion errors or generate less stable predictions over time.

We further evaluate the stability of our model under a longer extrapolation setting by extending the prediction interval from $[0.75, 1.0]$ to $[0.75, 1.5]$.
Since ground-truth frames are unavailable for the extended interval $[1.0, 1.5]$, we only provide qualitative results.
As shown in Figure~\ref{fig:longer_extrapolation}, our method maintains stable and continuous predictions over the extended horizon.
\subsection{Additional Quantitative Results}
\label{app:quantitative}
We report additional quantitative results on the MultiPart dataset and the FreeGave-GoPro dataset for the interpolation task.
As shown in Table~\ref{add_interpolation_tab}, our framework consistently outperforms conventional reconstruction-based methods, demonstrating its effectiveness in recovering temporally continuous appearance and motion under challenging dynamic scenes.

\begin{figure*}[t]
  \centering
  \includegraphics[width=5.4in]{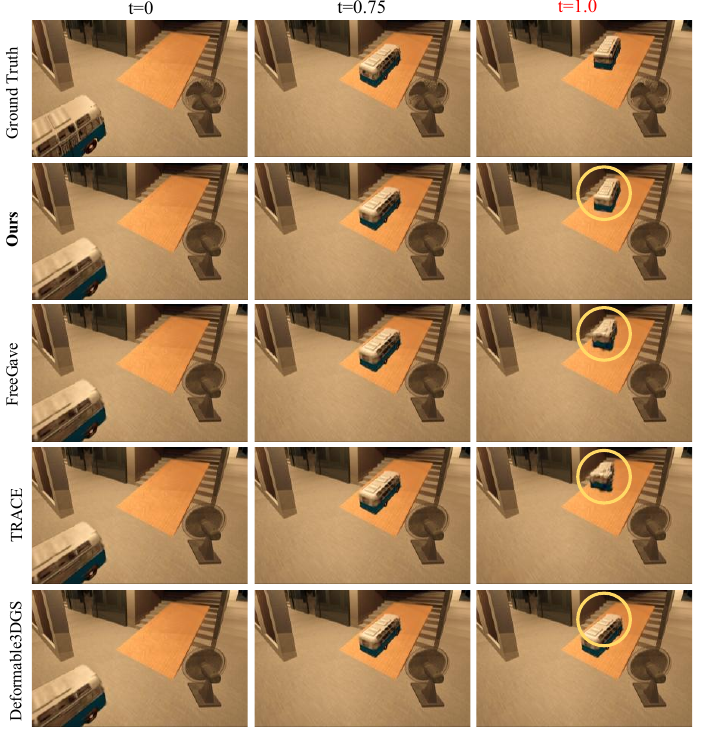}
  \caption{Qualitative Results  for Extrapolation of Bus Scene on PhysInOne Dataset.}
  \label{Bus}
  \vspace{-1.3em}
\end{figure*}
\begin{figure*}[t]
  \centering
  \includegraphics[width=5.4in]{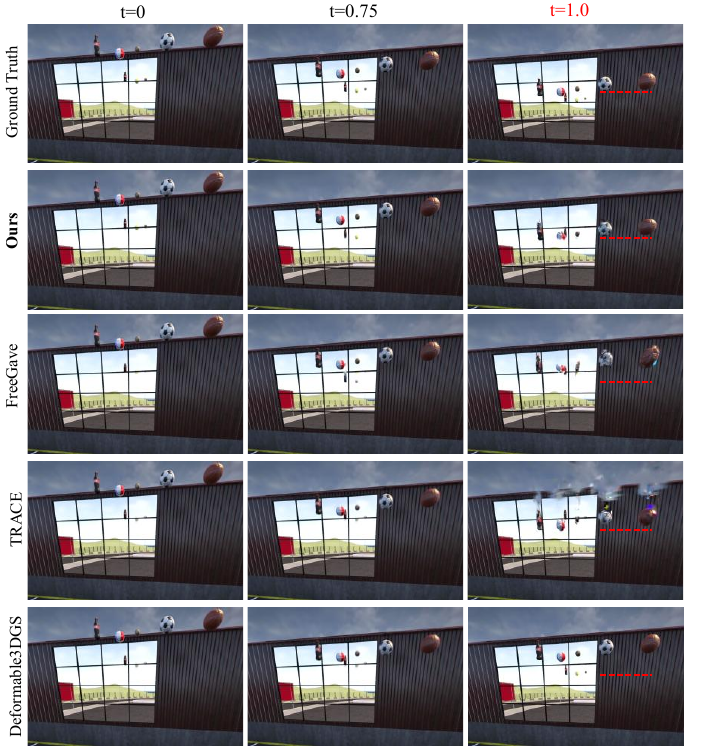}
  \caption{Qualitative Results  for Extrapolation of Freefall Scene on PhysInOne Dataset.}
  \vspace{-1.3em}
  \label{Freefall}
\end{figure*}
\begin{figure*}[t]
  \centering
  \includegraphics[width=5.4in]{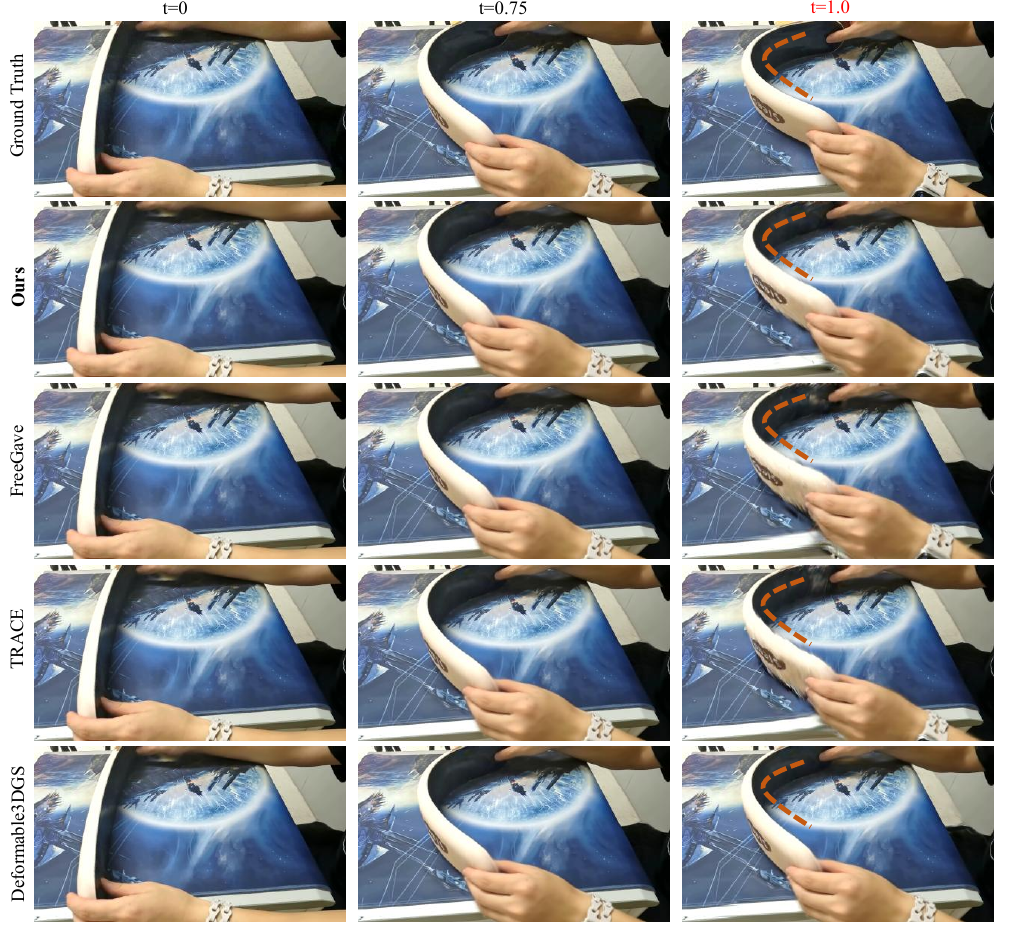}
  \caption{Qualitative Results  for Extrapolation of Wristrest Scene on FreeGave-GoPro Dataset.}
   \vspace{-1.3em}
  \label{Writrest}
\end{figure*}
\begin{figure*}[t]
  \centering
  \includegraphics[width=5.5in]{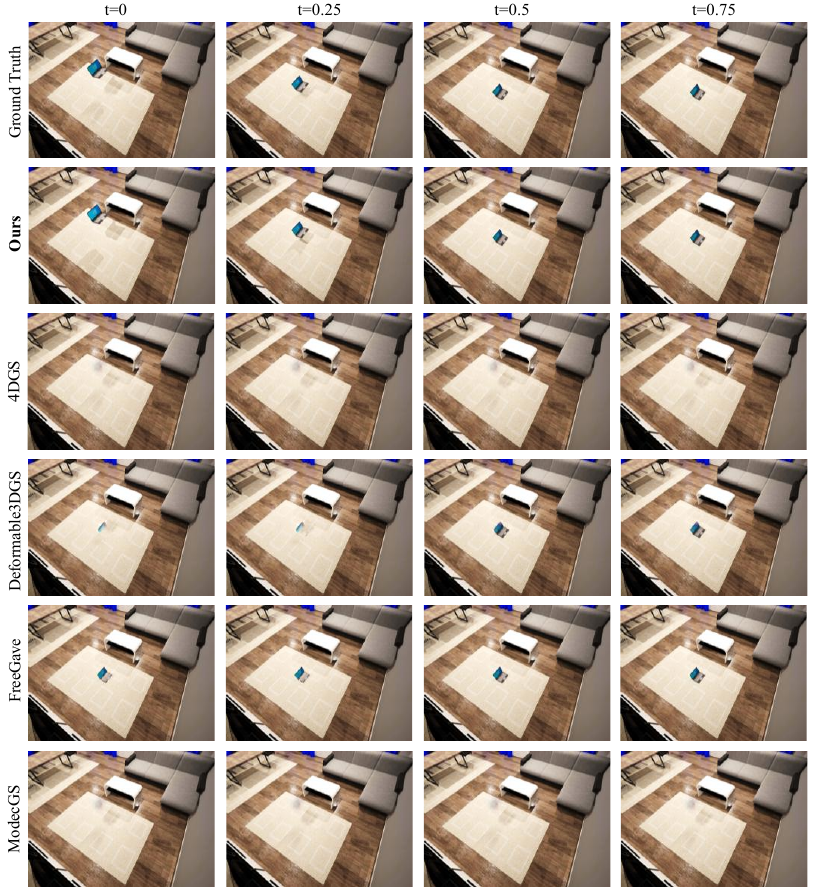}
  \caption{Qualitative Results  for Interpolation of Laptop Scene on PhysInOne Dataset.}
  \label{laptop}
\end{figure*}
\begin{figure*}[t]
  \centering
  \includegraphics[width=5.6in]{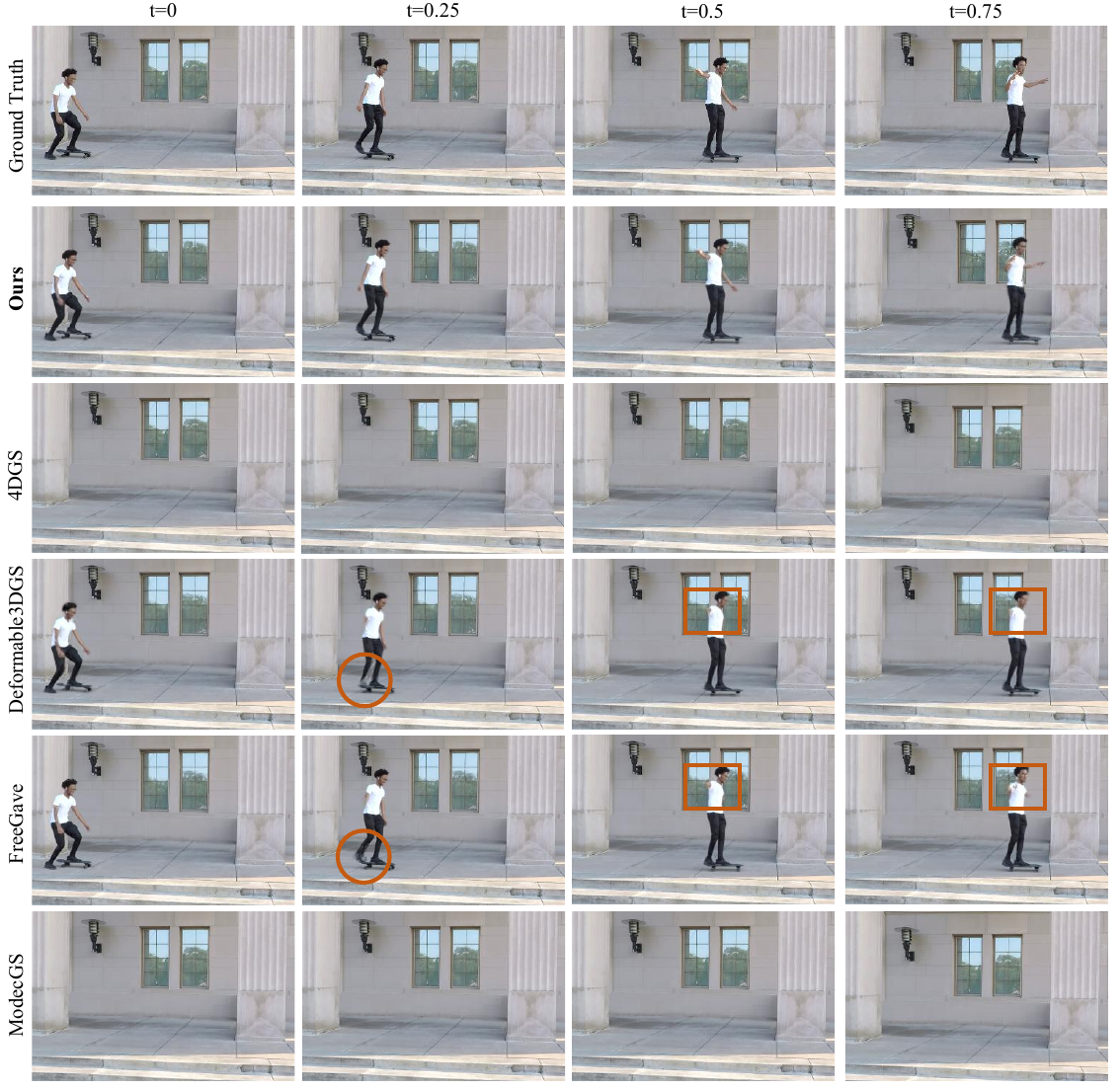}
  \caption{Qualitative Results  for Interpolation of Skating Scene on NVIDIA Dynamic Scene Dataset.}
  \label{skating}
\end{figure*}
\begin{figure}[h!]
  \centering
  \includegraphics[width=5.6in]{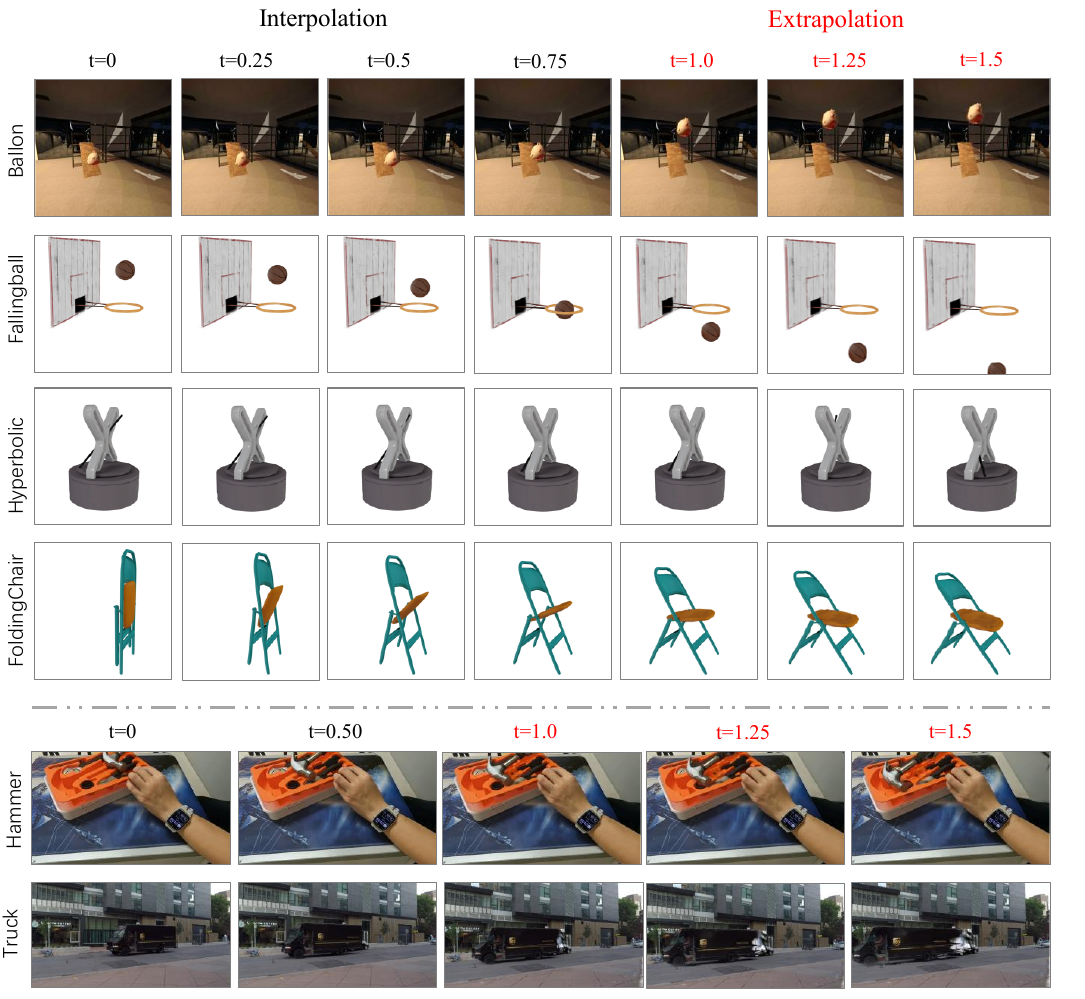}
  \caption{Qualitative Results of Longer Extrapolation on Four Datasets.}
  \label{fig:longer_extrapolation}
\end{figure}
\subsection{Details of All Datasets.}
\label{app:datasets}
\paragraph{FreeGave-GoPro Dataset.}
This dataset contains 6 challenging real-world dynamic scenes:
\begin{itemize}[topsep=2pt, itemsep=1pt, parsep=0pt, partopsep=0pt]
    \item \textbf{Pen \& Tape.}
    A person holds a pen and attempts to pass it through the hole of a static tape ring.
    Scene 2 includes more static clutter, introducing stronger occlusions and requiring more accurate separation between moving and static regions.
    The key challenge lies in forecasting an interaction where one object is expected to penetrate through another in the future.
    
    \item \textbf{Box.}
    A drawer-like box is pushed toward closure.
    The challenge comes from the tight coupling between the moving part and the static structure, which becomes particularly difficult under future extrapolation.

    \item \textbf{Hammer.}
    A hammer moves on top of a box.
    The scene involves direct contact between dynamic and static objects, requiring sharp separation of diverse motion patterns to preserve correct static/moving states in future frames.

    \item \textbf{Collision.}
    A cube and a cup move toward each other with different motion directions.
    The challenge is to maintain accurate object shapes and trajectories under future prediction.

    \item \textbf{Wrist Rest}
    A person bends a wrist rest.
    The object is deformable, and the motion is neither rigid nor purely part-wise rigid.
\end{itemize}
\textbf{Dynamic Multipart Dataset.}
This dataset features multiple articulated/deformable objects with part-wise motions:
\begin{itemize}[topsep=2pt, itemsep=1pt, parsep=0pt, partopsep=0pt]

    \item \textbf{Folding Chair.}
    A folding chair composed of three parts is unfolded.
    Different parts undergo distinct rotations.

    \item \textbf{Hyperbolic Slot.}
    A stick rotates through a hyperbolic-shaped slot.
    Only the stick is dynamic, making motion extrapolation challenging due to the complex static geometry.

    \item \textbf{Satellite.}
    A satellite with two wing doors and one main door opening, where different parts rotate in different directions.

    \item \textbf{Stove.}
    A home stove whose motion mainly involves closing its top cover plate.
\end{itemize}
\textbf{NVIDIA Dynamic Scenes Dataset.}
This dataset contains 2 real-world dynamic scenes:
\begin{itemize}[topsep=2pt, itemsep=1pt, parsep=0pt, partopsep=0pt]
    \item \textbf{Skating.}
    A person rides a skateboard while maintaining balance, and then makes a ``V'' sign on the skateboard.

    \item \textbf{Truck.}
    A truck drives along the street.
\end{itemize}
\textbf{PhysInone Dataset.}
We select 6 synthetic scenes governed by four physical effects (gravity, buoyancy, wind force, and friction):
\begin{itemize}[topsep=2pt, itemsep=1pt, parsep=0pt, partopsep=0pt]
    \item \textbf{Laptop.}
    A laptop falls freely in a room, lands on the floor, and then comes to rest.

    \item \textbf{Ballon.}
    The white rabbit-shaped balloon floats upward, slowly rising due to buoyancy.

    \item \textbf{Freefall.}
    A cola bottle, a baseball, a soccer ball, a basketball, and a volleyball fall side by side from a rooftop under gravity.

    \item \textbf{Slope.}
    A wooden plank forms an incline.
    A W-shaped balloon rises, hits the plank, and slides up it, pulling a rope tied to a bottle.
    Meanwhile, a bell pepper, two apples, a chips tube, and a lollipop roll down the plank under gravity onto the floor near a carpet and chair.

    \item \textbf{Bus.}
    A blue toy bus sits on a wooden ramp.
    A front fan pushes it up the incline, while a side fan nudges it sideways, changing its path until it stops mid-ramp.

    \item \textbf{Sandwich.}
    Two sandwichs slide on two different smooth surfaces under a fan’s airflow.
\end{itemize}
 \vspace{-1.0em}
\subsection{Training and Evaluation Resources}
\label{app:resource}
As the complexity of different scenes varies, the total number of Gaussians optimized per scene ranges from $30$k to $2$M.
Overall, our training speed is comparable to FreeGave, running at about $0.98\times$ the speed of FreeGave.
This overhead mainly comes from  additional training objectives, while the added network components remain lightweight.
At inference time, our rendering speed is comparable to Deformable3DGS.
All models are trained on a single NVIDIA RTX 3090 GPU.

\vspace{-0.6em}
\subsection{Limitation of Our Model}
\label{limitation}
Our method remains challenged by highly deformable scenes, such as Newtonian fluids, due to the inherent representation limitations shared by Gaussian Splatting-based methods.
In such scenarios, the scene geometry undergoes continuous and complex deformation, making it difficult for discrete Gaussian primitives to faithfully capture fine-grained topology and appearance changes.
Moreover, PhyMo captures physical dynamics at the kinematic level without explicitly modeling forces, mass, or material parameters, which may limit its ability to handle complex physical interactions.

\clearpage
\section*{NeurIPS Paper Checklist}

\begin{enumerate}

\item {\bf Claims}
    \item[] Question: Do the main claims made in the abstract and introduction accurately reflect the paper's contributions and scope?
    \item[] Answer: \answerYes{} 
    \item[] Justification: Our papers main contribution has been clearly declared in the abstract and introduction.

\item {\bf Limitations}
    \item[] Question: Does the paper discuss the limitations of the work performed by the authors?
    \item[] Answer: \answerYes{} 
    \item[] Justification: We include the limitation part in the Appendix.

\item {\bf Theory assumptions and proofs}
    \item[] Question: For each theoretical result, does the paper provide the full set of assumptions and a complete (and correct) proof?
    \item[] Answer: \answerNA{} 
    \item[] Justification: The paper does not include theorectical results.

    \item {\bf Experimental result reproducibility}
    \item[] Question: Does the paper fully disclose all the information needed to reproduce the main experimental results of the paper to the extent that it affects the main claims and/or conclusions of the paper (regardless of whether the code and data are provided or not)?
    \item[] Answer: \answerYes{} 
    \item[] Justification: A detailed description of our method is provided, and the code will be released upon acceptance.

\item {\bf Open access to data and code}
    \item[] Question: Does the paper provide open access to the data and code, with sufficient instructions to faithfully reproduce the main experimental results, as described in supplemental material?
    \item[] Answer: \answerNo{} 
    \item[] Justification: The code is not included for now. It will be released upon acceptance.

\item {\bf Experimental setting/details}
    \item[] Question: Does the paper specify all the training and test details (e.g., data splits, hyperparameters, how they were chosen, type of optimizer) necessary to understand the results?
    \item[] Answer: \answerYes{} 
    \item[] Justification: We include them in the Experiment Section and also more details in the Appendix.

\item {\bf Experiment statistical significance}
    \item[] Question: Does the paper report error bars suitably and correctly defined or other appropriate information about the statistical significance of the experiments?
    \item[] Answer: \answerYes{} 
    \item[] Justification: We perform the experiments multiple times and compute the average.

\item {\bf Experiments compute resources}
    \item[] Question: For each experiment, does the paper provide sufficient information on the computer resources (type of compute workers, memory, time of execution) needed to reproduce the experiments?
    \item[] Answer: \answerYes{} 
    \item[] Justification: We provide sufficient information on the computer resources for experiments in the Appendix.
    
\item {\bf Code of ethics}
    \item[] Question: Does the research conducted in the paper conform, in every respect, with the NeurIPS Code of Ethics \url{https://neurips.cc/public/EthicsGuidelines}?
    \item[] Answer: \answerYes{} 
    \item[] Justification: We follow the NeurIPS Code of Ethics.

\item {\bf Broader impacts}
    \item[] Question: Does the paper discuss both potential positive societal impacts and negative societal impacts of the work performed?
    \item[] Answer: \answerYes{} 
    \item[] Justification: We discussed the potential positive impact of our work, as we outlined some of its application in the introduction section of our main text. As for negative social impacts, we have not yet identified significant negative implications.
    
\item {\bf Safeguards}
    \item[] Question: Does the paper describe safeguards that have been put in place for responsible release of data or models that have a high risk for misuse (e.g., pre-trained language models, image generators, or scraped datasets)?
    \item[] Answer: \answerNA{} 
    \item[] Justification: The paper poses no such risks.

\item {\bf Licenses for existing assets}
    \item[] Question: Are the creators or original owners of assets (e.g., code, data, models), used in the paper, properly credited and are the license and terms of use explicitly mentioned and properly respected?
    \item[] Answer: \answerYes{} 
    \item[] Justification: The code we use has been properly cited in the paper.

\item {\bf New assets}
    \item[] Question: Are new assets introduced in the paper well documented and is the documentation provided alongside the assets?
    \item[] Answer: \answerNA{} 
    \item[] Justification: The paper does not release new datasets or new assets.

\item {\bf Crowdsourcing and research with human subjects}
    \item[] Question: For crowdsourcing experiments and research with human subjects, does the paper include the full text of instructions given to participants and screenshots, if applicable, as well as details about compensation (if any)? 
    \item[] Answer: \answerNA{} 
    \item[] Justification: The paper does not involve crowdsourcing nor research with human subjects.

\item {\bf Institutional review board (IRB) approvals or equivalent for research with human subjects}
    \item[] Question: Does the paper describe potential risks incurred by study participants, whether such risks were disclosed to the subjects, and whether Institutional Review Board (IRB) approvals (or an equivalent approval/review based on the requirements of your country or institution) were obtained?
    \item[] Answer:  \answerNA{} 
    \item[] Justification: The paper does not involve crowdsourcing nor research with human subjects.

\item {\bf Declaration of LLM usage}
    \item[] Question: Does the paper describe the usage of LLMs if it is an important, original, or non-standard component of the core methods in this research? Note that if the LLM is used only for writing, editing, or formatting purposes and does \emph{not} impact the core methodology, scientific rigor, or originality of the research, declaration is not required.
    \item[] Answer: \answerNA{} 
    \item[] Justification: LLM is used only for writing, editing, the core method development in this paper does not involve LLMs as any important components.

\end{enumerate}

\end{document}